\documentclass[runningheads]{llncs}

\usepackage{eccv}
\RequirePackage[width=122mm,left=12mm,paperwidth=146mm,height=193mm,top=12mm,paperheight=217mm]{geometry}

\usepackage{relsize}
\usepackage{wrapfig}

\usepackage{eccvabbrv}

\usepackage{graphicx}
\usepackage{booktabs}

\usepackage[accsupp]{axessibility}  

\usepackage{hyperref}

\usepackage{orcidlink}

\usepackage[inline]{enumitem} 
\usepackage{color}
\usepackage{multirow}
\usepackage[makeroom]{cancel}
\usepackage{bm}

\definecolor{turquoise}{cmyk}{0.65,0,0.1,0.3}
\definecolor{purple}{rgb}{0.65,0,0.65}
\definecolor{dark_purple}{rgb}{0.5,0,0.5}
\definecolor{dark_green}{rgb}{0, 0.5, 0}
\definecolor{orange}{rgb}{0.8, 0.6, 0.2}
\definecolor{red}{rgb}{0.8, 0.2, 0.2}
\definecolor{darkred}{rgb}{0.6, 0.1, 0.05}
\definecolor{blueish}{rgb}{0.0, 0.3, .6}
\definecolor{light_gray}{rgb}{0.7, 0.7, .7}
\definecolor{pink}{rgb}{1, 0, 1}
\definecolor{greyblue}{rgb}{0.25, 0.25, 1}

\newcommand{\qm}[1]{{\color{black}#1}}

\newcommand{\ApproachName}{DECOLLAGE\xspace}

\newcommand{\StopGradient}{\cancel\nabla}

\newcommand{\smallcheck}{$({\color{blue}\surd}\,)$}
\newcommand{\smallxmark}{$(\mathlarger{\mathlarger{\mathlarger{\times}}})$}
\begin{document}

\title{DECOLLAGE: 3D Detailization by Controllable, Localized, and Learned Geometry Enhancement} 

\titlerunning{DECOLLAGE}

\author{Qimin Chen\inst{1, 2}\orcidlink{0009-0004-8447-0137} \and
Zhiqin Chen\inst{2}\orcidlink{0000-0001-7835-1618} \and
Vladimir G. Kim\inst{2}\orcidlink{0000-0002-3996-6588} \and Noam Aigerman\inst{3}\orcidlink{0000-0002-9116-4662} \and \\ Hao Zhang\inst{1, 4}\orcidlink{0000-0003-1991-119X} \and Siddhartha Chaudhuri\inst{2}\orcidlink{0009-0009-8588-1436}}

\authorrunning{Q.~Chen et al.}


\institute{Simon Fraser University \and Adobe Research \and University of Montreal \and Amazon}

\maketitle

\begin{abstract}
    We present a 3D modeling method which enables end-users to refine or {\em detailize\/} 3D shapes using machine learning, expanding the capabilities of AI-assisted 3D content creation. Given a coarse voxel shape (e.g., one produced with a simple box extrusion tool or via generative modeling), a user can directly ``paint'' desired target styles representing compelling geometric details, from input exemplar shapes, over different regions of the coarse shape. These regions are then up-sampled into high-resolution geometries which adhere with the painted styles. To achieve such controllable and localized 3D detailization, we build on top of a Pyramid GAN by making it {\em masking-aware\/}. We devise novel structural losses and priors to ensure that our method preserves both desired coarse structures and fine-grained features even if the painted styles are borrowed from diverse sources, e.g., different semantic parts and even different shape categories. Through extensive experiments, we show that our ability to localize details enables novel interactive creative workflows and applications. Our experiments further demonstrate that in comparison to prior techniques built on global detailization, our method generates structure-preserving, high-resolution stylized geometries with more coherent shape details and style transitions.
    \keywords{3D detailization \and Controllable 3D generation \and Generative adversarial network \and High-resolution geometry}
\end{abstract}
\section{Introduction}
\label{sec:intro}


%

Customized 3D content is becoming more widely available, driven by rapid advances in generative AI and increasing demand from 
computer games, AR/VR, and e-commerce. Recently, deep generative models based on diffusion and vision-language models have made significant waves in improving the accessibility (e.g., via text prompting) and ingenuity of generated content, as well as enabling zero-shot learning. However, while effective at creating coarse content, the latest methods along these fronts, 
e.g.,~\cite{lin2023magic3d,poole2022dreamfusion,gao2022get3d,tang2023makeit3d}, still lack the ability to generate and precisely control high-quality {\em geometric details}. Also, their slow speed remains a roadblock to integrating them into artists' conventional workflows.

\begin{figure}[t]
  \includegraphics[width=\textwidth]{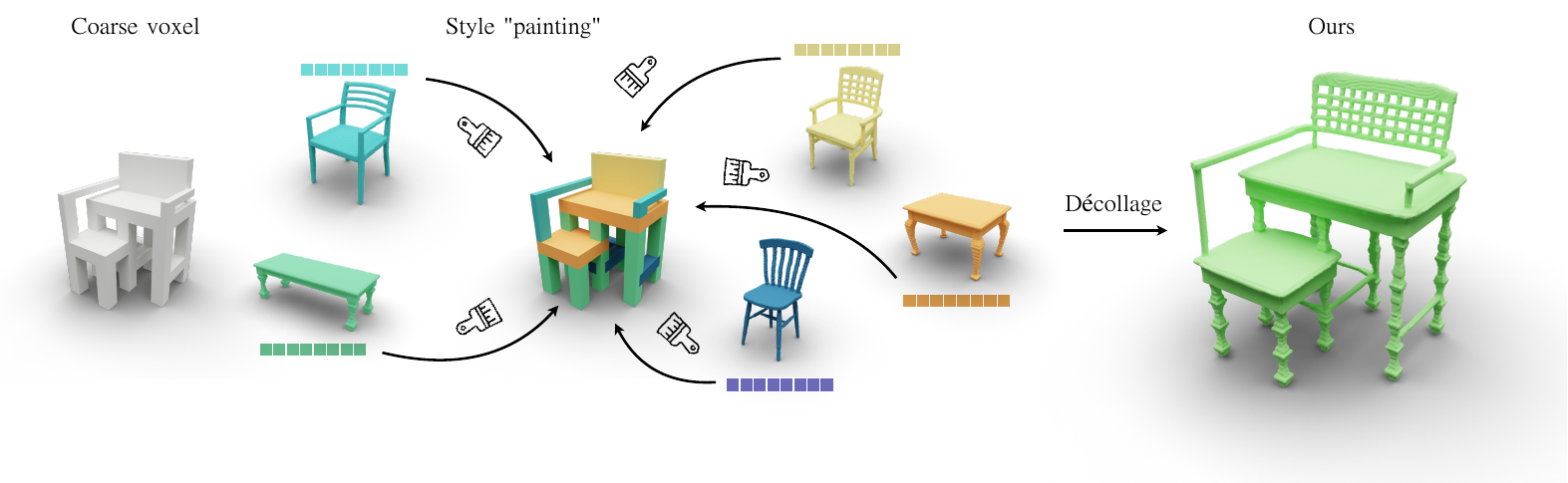}
  \vspace{-11mm}
  \caption{D\'{e}collage is an art form created by ``cutting/removing pieces of an original image''\textsuperscript{1}. When ``painting'' a style exemplar with geometric details over a region of a coarse shape, coarse surfaces are removed to unveil a {\em detailized\/} version to mimic the exemplar. We show an out-of-distribution chair-like shape detailized via {\em style mixing\/}, where five exemplars ``d\'{e}collaged" the coarse voxels.}
  \label{fig:teaser}
  \vspace{-3mm}
\end{figure}

In this paper, we propose a learning-based method that enables novice users to add geometric details to a coarse 3D shape by selecting regions on it and assigning them the {\em styles} of exemplar shapes with compelling geometric details. Each region is upsampled and {\em detailized} by a neural network to replicate the corresponding exemplar's detail style, while preserving the overall structure of the coarse input {\em content} shape. In general, when detailizing multiple regions using separate and possibly diverse style exemplars, i.e., ``style mixing,'' as shown in Figure~\ref{fig:teaser}, the goal of our region-specific, {\em localized} 3D detailization is to produce structure-preserving and globally coherent results in terms of shape details and part connections, across feature scales and shape categories. Style mixing from different shape categories offers additional design freedom which can boost the creative potential of the generated shapes without compromising structural validity and functionality of the input content shapes, as shown in Figure~\ref{fig:teaser}.

\footnotetext[1]{{\protect\url{https://en.wikipedia.org/wiki/Decollage}}}

\begin{figure}[t]
\vspace{-3mm}
\begin{minipage}[b]{.48\textwidth}
\begin{picture}(0, 95)
  \put(0, 0){\includegraphics[width=\textwidth]{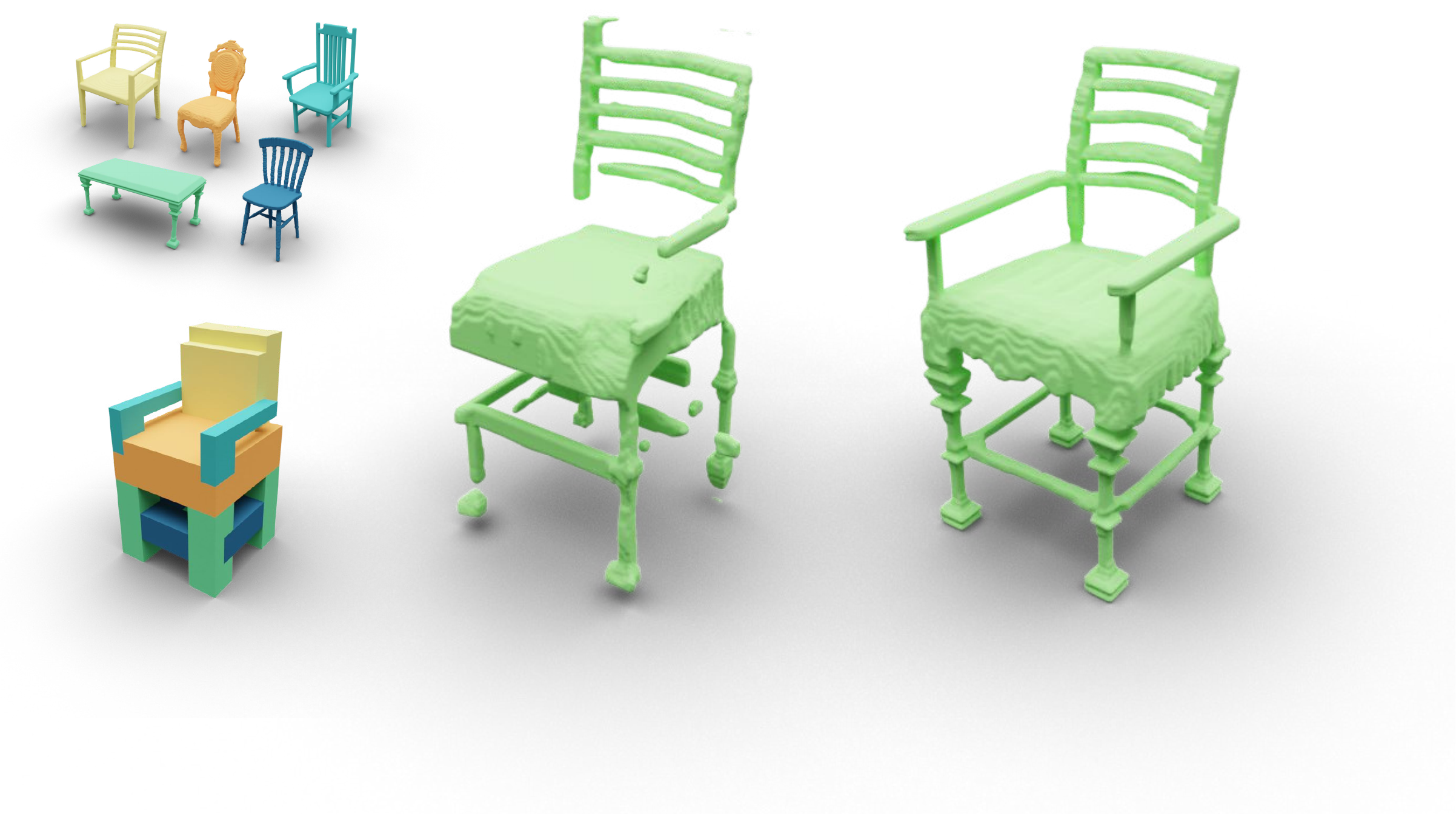}}
  \put(10, 15){\scriptsize Content}
  \put(15, 60){\scriptsize Styles}

  \put(45, 15){\scriptsize (a) DECOR-GAN$^{*}$}
  \put(115, 15){\scriptsize (b) Ours}
\end{picture}
  \vspace{-5mm}
  \caption{DECOR-GAN$^{*}$~\cite{chen2021decor} (a) with na\"ive local controllability generates disconnected structures and floating pieces. Our method DECOLLAGE (b) fares much better in preserving global structure and generating local geometric details.}
  \label{fig:seg_decorgan}
\end{minipage}
\hspace{2mm}
\begin{minipage}[b]{.49\textwidth}
\begin{picture}(0, 95)
  \put(0, 0){\includegraphics[width=\textwidth]{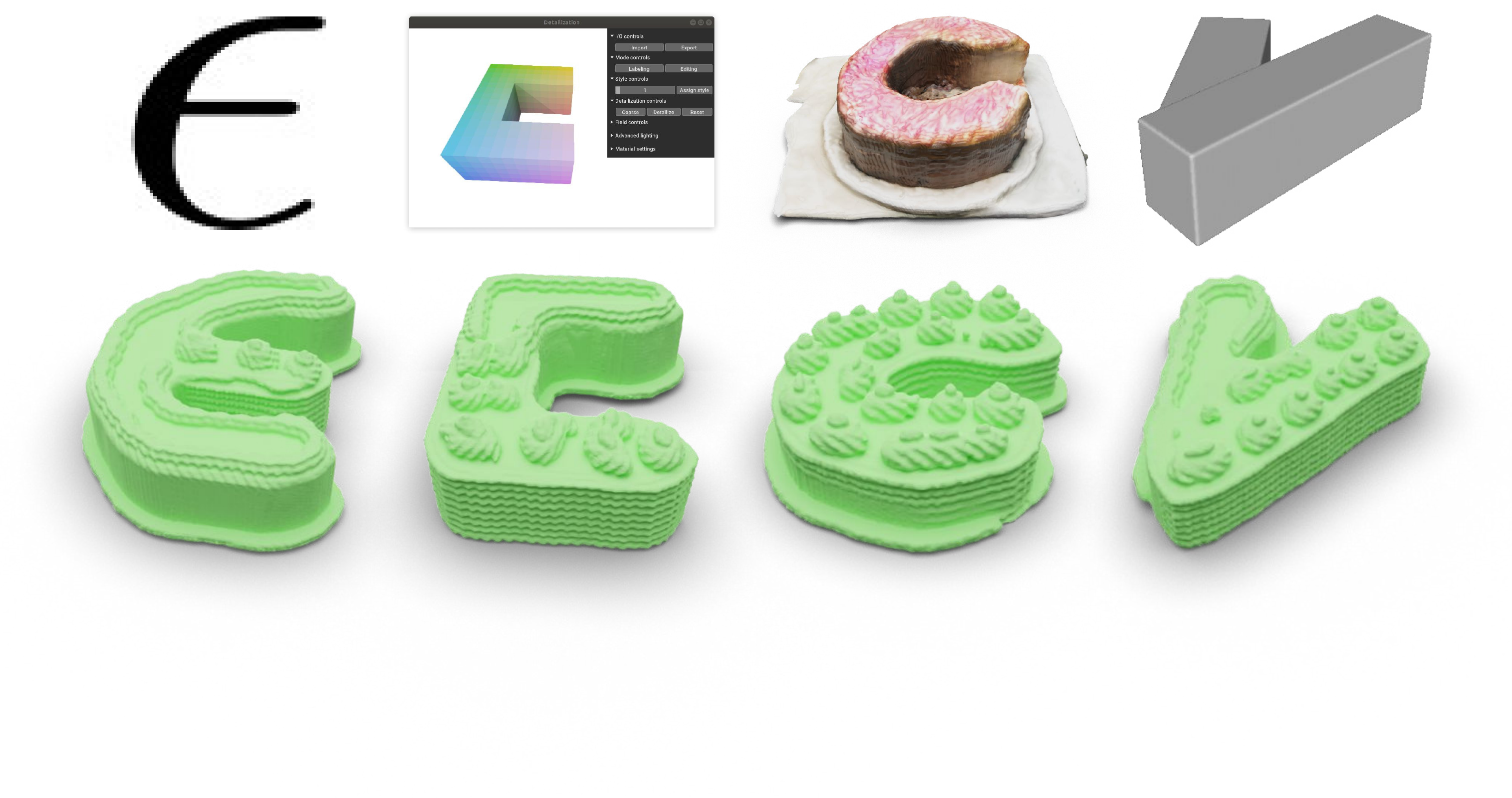}}
  \put(20, 15){\scriptsize (a)}
  \put(60, 15){\scriptsize (b)}
  \put(100, 15){\scriptsize (c)}
  \put(140, 15){\scriptsize (d)}
\end{picture}
  \vspace{-5mm}
  \caption{Application: detailizing shapes from various sources including (a) extruding 2D profiles; (b) coarse voxels created via an interactive user interface (see supplementary); (c) shapes generated by text-to-3D model; (d) simple CAD primitives.}
  \label{fig:application}
\end{minipage}
\vspace{-5mm}
\end{figure}

Enabling localized style control via exemplar shapes is a natural content creation paradigm, which addresses both the questions of {\em which} detail to generate and {\em where} to generate. However, the problem is technically challenging, especially with style mixing. Even when both the content and style shapes happen to be semantically segmented, decoupled assignment of details to target areas inevitably leads to structural inconsistencies, especially over joint regions, as the details may not trivially mix.
Besides requiring special treatment to ensure coherent part connections, the network also needs to have a global understanding of the whole shape while applying detail locally. Prior methods for conditional detailization~\cite{chen2021decor,chen2023shaddr} are designed to deal with a single style exemplar and do not perform well when there is significant structural dissimilarity to the 
content shape.
In addition, many style configurations may have never been observed during training, leading to out-of-distribution failures: see Figure~\ref{fig:seg_decorgan}.
%

%

To address the above challenges, our method leverages a 
{\em hierarchical} backbone architecture for generative adversarial learning, i.e., a Pyramid GAN~\cite{denton2015deep,wu2022learning,wang2022singrav,karnewar_3InGan_3dv_22}. This enables our network to capture both global structures using coarse-level reasoning, as well as local geometric details at finer levels. Accordingly, our network is trained with both a global discriminator and a local, style-conditioned one, where we employ an {\em adaptive $\alpha$ weighting\/} to adjust the importance of the discriminators depending on proximity to style transition regions. In addition, we propose novel network losses to encourage structure preservation during upsampling. Specifically, we ensure that the coarse shape is preserved under resampling at different resolutions by both downsampling and upsampling it.
%
%
Finally, since local style control requires the coarse targets to be partitioned into regions to guide the generation, we propose several data augmentation mechanisms to generate segmented coarse targets from detailed sources during self-supervised training, by randomly changing their part structures, scales, and orientations.


In summary, our work offers the first method for {\em interactive\/}, {\em controllable\/}, and {\em localized\/} geometry detail generation, unlike existing shape detalization works~\cite{chen2021decor,chen2023shaddr} which only offer global style control. On the technical front, the adaptive $\alpha$ weighting between discriminators has not been used in prior works, and is essential to our interactive workflow. Although Pyramid GAN has been heavily explored before, it alone is insufficient to tackle the challenge of style mixing. To this end, our key contribution is the use of novel structure-preserving losses tailored for the Pyramid architecture with discriminator adaptivity to overcome incoherent structures.
Once trained, our network enables novel {\em interactive} 3D modeling, allowing both structure editing by end-users and automated per-region detailization, as shown in Figure \ref{fig:application}. This application also showcases the versatility of our modeling paradigm through DECOLLAGE, in terms of 3D content shapes.
%

%
%
%
We conduct experiments to show that our approach performs significantly better than relevant baselines on 3D detailization by borrowing details across different categories of shapes. We further demonstrate that our method outperforms prior works on tasks they were designed to handle, i.e., a standard example-based detailization with a \emph{single} style, from the same category~\cite{chen2021decor}. 
Lastly, we showcase applications of our method to enable various workflows, such as creating detailed shapes from coarse labeled blocks, and detailizing coarse generated shapes by painting style labels.


\section{Related work}
\label{sec:related}

While our approach might appear to tackle a similar problem to 3D voxel up-sampling~\cite{peng2020convolutional,chibane2020implicit,siddiqui2021retrievalfuse,chen2021nmc,chen2022ndc,sellan2023reach}, however, a critical distinction is that it aims to generate new features and details. We thus review prior 3D generative models and shape detailization techniques. 


\vspace{3mm} \noindent
\textbf{3D generative models.}
Various 3D generative models have been introduced for point clouds~\cite{achlioptas2018learning,zeng2022lion}, voxels~\cite{choy20163d,wu2016learning}, neural implicit functions~\cite{chen2019imnet,mescheder2019occupancy,park2019deepsdf}, neural radiance fields~\cite{mildenhall2021nerf,poole2022dreamfusion,lin2023magic3d}, and hybrid representations~\cite{li2023diffusion,hui2022neural,gao2022get3d}. These approaches are predominantly empowered by variational autoencoders (VAEs)~\cite{kingma2013auto}, generative adversarial networks (GANs)~\cite{goodfellow2020generative}, or diffusion probabilistic models~\cite{sohl2015deep,ho2020denoising}.

Despite significant progress in this field, very few works offer controllable and interactive 3D shape generation for modeling applications. Notably, Point-E~\cite{nichol2022point}, Shap-E~\cite{jun2023shap}, and One-2-3-45~\cite{liu2023one} can generate a 3D model from text or single image inputs, with processing times ranging from 30 seconds to over a minute. DECOR-GAN~\cite{chen2021decor} and ShaDDR~\cite{chen2023shaddr} can efficiently produce a detailed 3D shape from coarse voxel inputs, taking less than 1 to 2 seconds. Although ShaDDR provides interactivity during generation, it only offers global style control. Our work builds upon DECOR-GAN to deliver the first \textit{controllable} and \textit{localized} interactive modeling experience, while introducing new formulations and technical contributions that enable local style control and improve robustness in handling coarser and out-of-distribution inputs.

\vspace{3mm} \noindent
\textbf{3D shape detailization.}
Apart from classic methods that apply displacement maps or volumetric textures~\cite{kajiya1989rendering,neyret1998modeling} on the surfaces to represent geometric details, recent methods have been proposed to perform local geometric operations on a coarse mesh to synthesize surface details.
Berkiten et al.~\cite{berkiten2017learning} employs metric learning to transfer displacement maps from a high-quality 3D mesh to a coarse mesh.
Hertz et al.~\cite{hertz2020deep} learns geometric texture from a single reference 3D mesh and is able to apply the learned texture to a new shape.
Neural Subdivision~\cite{liu2020neural} also learns local geometric features and is able to transfer them via mesh subdivision.
Leveraging differentiable rendering of meshes, Paparazzi~\cite{liu2018paparazzi} and Text2Mesh~\cite{michel2022text2mesh} can generate geometric details on the mesh surface conditioned on the style of a reference image or input text, respectively.
\qm{3DStyleNet~\cite{yin2021_3DStyleNet} transfers geometric and texture styles from one shape to another.}
However, none of the aforementioned methods is capable of altering the topology of the coarse mesh, therefore restricting the range of geometric styles they can synthesize.

Other methods aim to synthesize geometric details by replicating local patches from a reference shape, thereby overcoming the topology constraint.
Inspired by image quilting~\cite{efros2001image}, mesh quilting~\cite{zhou2006mesh} can detailize the surface of a coarse mesh by copying, deforming, and stitching local patches of a given geometric texture patch.
SketchPatch~\cite{fish2020sketchpatch} adopts a PatchGAN~\cite{isola2017image} discriminator to mimic the local style of a reference image in order to stylize plain solid-lined sketches.
DECOR-GAN~\cite{chen2021decor} similarly utilizes PatchGAN for generating detailed voxel shapes from input coarse voxels, with the geometric style of the generated shape copied from one detailed reference voxel model.
ShaDDR~\cite{chen2023shaddr} improves the generated geometry of \cite{chen2021decor} by leveraging a 2-level hierarchical GAN, and introduces texture generation.
DMTET~\cite{shen2021dmtet} can detailize a coarse voxel shape into a detailed mesh through differentiable marching tetrahedra.
While the ability to generate arbitrary topology is desirable, it also comes with drawbacks.
For instance, DECOR-GAN can synthesize impressive patterns, yet it is prone to producing disconnected parts and redundant floating pieces; see Figure~\ref{fig:seg_decorgan}. Our method effectively addresses this issue via novel structure-preserving losses and adaptive $\alpha$ weighting of style and global discriminators.

\vspace{3mm} \noindent
\textbf{Hierarchical GANs.}
The pyramid structure of our network architecture draws inspiration from hierarchical GANs~\cite{denton2015deep,wang2018high,zhang2017stackgan,karras2018progressive}, a structure widely employed in various tasks, including image~\cite{shaham2019singan} and 3D~\cite{wu2022learning,wang2022singrav,karnewar_3InGan_3dv_22} generation.
Multi-level generation based on different scales has been applied outside the scope of GANs to facilitate generation of structurally diverse shapes~\cite{Li_2023_CVPR}.
We apply hierarchical GAN and devise novel loss functions to generate shapes that are both globally plausible and locally detailed.

\section{Method}
\label{sec:method}

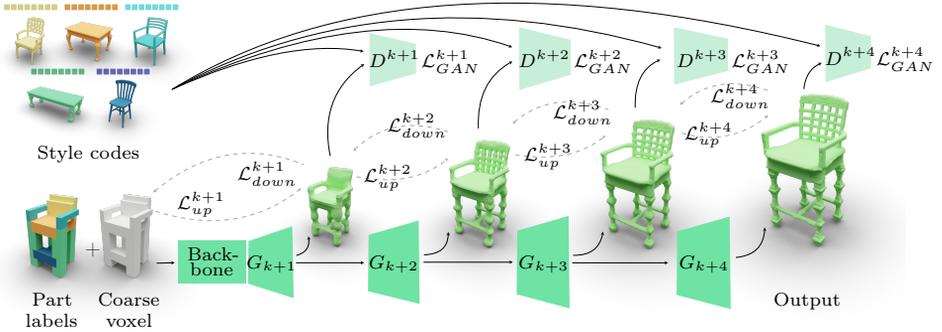
\begin{figure*}[t]
\begin{picture}(510, 125)
\centering
  \put(0, 0){\includegraphics[width=1.0\linewidth]{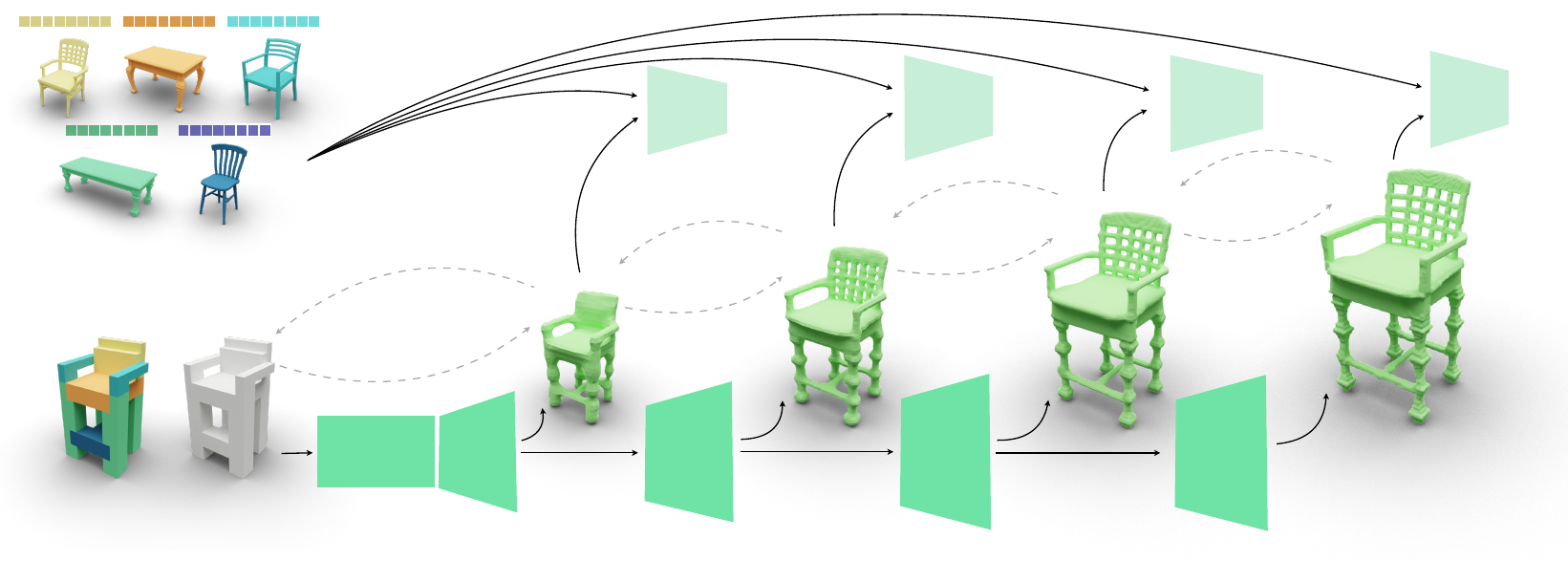}}
  \put(15, 8){\scriptsize Part}
  \put(12.5, 0){\scriptsize labels}
  \put(40, 8){\scriptsize Coarse}
  \put(43, 0){\scriptsize voxel}
  \put(35, 27){\scriptsize +}
  \put(17, 63){\scriptsize Style codes}

  \put(74, 25){\scriptsize Back-}
  \put(75, 19){\scriptsize bone}
  \put(96.5, 22){\scriptsize $G_{k+1}$}
  \put(143, 22){\scriptsize $G_{k+2}$}
  \put(200, 22){\scriptsize $G_{k+3}$}
  \put(261, 22){\scriptsize $G_{k+4}$}

  \put(297, 8){\scriptsize Output}

  \put(93, 55){\scriptsize $\mathcal{L}_{down}^{k+1}$}
  \put(70, 44){\scriptsize $\mathcal{L}_{up}^{k+1}$}
  
  \put(150, 73){\scriptsize $\mathcal{L}_{down}^{k+2}$}
  \put(141, 54){\scriptsize $\mathcal{L}_{up}^{k+2}$}
  
  \put(213, 78){\scriptsize $\mathcal{L}_{down}^{k+3}$}
  \put(202, 62){\scriptsize $\mathcal{L}_{up}^{k+3}$}
  
  \put(273, 84){\scriptsize $\mathcal{L}_{down}^{k+4}$}
  \put(263, 70){\scriptsize $\mathcal{L}_{up}^{k+4}$}

  \put(143, 97){\scriptsize $D^{k+1}$}
  \put(163, 97){\scriptsize $\mathcal{L}_{GAN}^{k+1}$}
  \put(200, 97){\scriptsize $D^{k+2}$}
  \put(221, 97){\scriptsize $\mathcal{L}_{GAN}^{k+2}$}
  \put(260, 97){\scriptsize $D^{k+3}$}
  \put(281, 97){\scriptsize $\mathcal{L}_{GAN}^{k+3}$}
  \put(316.5, 98){\scriptsize $D^{k+4}$}
  \put(336, 98){\scriptsize $\mathcal{L}_{GAN}^{k+4}$}
\end{picture}
\vspace{-4mm}
  \caption{Network architecture. Conditioned on a set of style codes associated with each segmented part, the network upsamples the coarse content voxel with part labels into detailed geometries in multiple resolutions. For each upsampling level $j$, the discriminator enforces the local patches of each part in the upsampled geometry to be plausible with respect to the styles they are conditioned on. The structure-preserving losses $\mathcal{L}_{down}^{j}$ and $\mathcal{L}_{up}^{j}$ enforce the structure of the output to be consistent with the input.
  }
  \label{fig:pipeline}
  \vspace{-3mm}
\end{figure*}

Our method requires only a few (e.g., $16$) detailed ``style'' shapes $\mathcal{S} = \{s_{1}, \dots, s_{N}\}$ with varying geometric styles as training data, where each style shape $s_i$ is a $2^K \times 2^K \times 2^K$ high-resolution voxel grid. We also assume all the shapes are co-segmented into meaningful parts, so that each voxel $v \in s_i$ is associated with a part label $P(v)$.
After training, given a $2^k \times 2^k \times 2^k$ coarse voxel grid $c$ where each voxel $v \in c$ is associated with a part label $P(v)$ and a style $S(v) \in \{1, \dots, N\}$, our model can generate a high-resolution, detailed shape that has the overall structure of the input $c$, while the local geometric detail in the region corresponded to a input voxel $v$ follows the style exhibited in $s_{S(v)}$.
In our experiments, we use $k=4$ and $K=8$.

In this section, we first introduce our network architecture - a pyramid GAN with $(K-k)$ levels, in Sec.~\ref{sec:network_architecture}.
Next, in Sec.~\ref{sec:loss_functions}, we apply masked adversarial losses to ensure that the generated geometry follows the styles specified by the input voxels, while an adaptive $\alpha$ weighting scheme is developed to promote proper connectivity in style transition regions.
We also devise two novel loss functions tailored for our pyramidal network architecture to preserve the structure of the input coarse voxels.
Our method requires coarse voxels $c$ with diverse structures and part segmentation during training. Therefore, in Sec.~\ref{sec:data_augmentation}, we propose a data augmentation technique to take full advantage of the $N$ segmented style shapes in $\mathcal{S}$ and use them to synthesis the coarse voxels for our training.

\subsection{Network architecture}
\label{sec:network_architecture}

Our network is shown in Figure~\ref{fig:pipeline}.
For each style shape $s_i$, we associate it with an optimizable 8-dimensional latent code $z_i$ to represent its geometric style; see Figure~\ref{fig:pipeline} top-left.
The input to our model is a coarse voxel grid $c$ of resolution $2^k$, where each voxel $v$ contains its binary occupancy $O(v)$, part label $P(v)$ and a latent code $z_{S(v)}$ corresponding to its designated style $S(v)$; see in Figure~\ref{fig:pipeline} bottom-left.
Our model contains a backbone and a pyramid of generator networks $G^j$ that upsample the input into occupancy voxels of size $2^j$, where $j \in \{k+1, \dots, K\}$. Each $G^j$ doubles the size of its input; see Figure~\ref{fig:pipeline} bottom. Both the backbone and the generators are 3D convolutional neural networks (CNNs).

Correspondingly, a pyramid of 3D CNN PatchGAN~\cite{isola2017image} discriminators $D^j$ are employed to ensure that the shape generated at each resolution level is plausible; see Figure~\ref{fig:pipeline} top-right.
Each $D^j$ inputs a generated occupancy grid at $2^j$ resolution and outputs a voxel grid of the same resolution, while each output voxel has $N+1$ channels. The first $N$ channels of an voxel $v$ in the discriminator output represent the likelihood that the local patch covered by $v$'s receptive field is from one of the $N$ style shapes, thus the first $N$ channels are {\em style-specific\/} discriminators. The $(N+1)$-th channel can be viewed as a {\em global\/} discriminator that evaluates the likelihood of the patch being plausible for a 3D shape. We denote the style-specific discriminators as $D^j_i$ for level $j$ and style $i$, and the global discriminator as $D^j_*$.

A main advantage of pyramid GANs is that we can have different receptive fields at different levels, so that the coarse levels only focus on generating coherent structures, while the fine levels pay more attention to generating plausible details. This provides better generalizability to inputs with drastically different structures compared to the training style shapes. We set the discriminator receptive fields to $7^3$ and $9^3$ for the first two levels and $18^3$ for the rest.

\subsection{Loss functions}
\label{sec:loss_functions}

\noindent
\textbf{Reconstruction loss.}
Given a high-resolution style shape $s_i$, we downsample it to a lower resolution and use it as an input coarse shape. We can directly apply a reconstruction loss as we have the ground truth (GT). Denote the downsampled shape at resolution $2^j$ as $s^j_i$, for each style $i$ and resolution level $j$, we have
\begin{align}
\mathcal{L}_{recon} = \mathbb{E}_{v} ( G^j(s^k_i)[v] - O(s^j_i[v]) )^2,
\end{align}
where $v$ iterates the indices of all voxels, $G^j(\cdot)$ is the output voxel grid of the generator, and $s[v]$ queries the voxel in $s$ at index $v$.
Note that the style $S(s^k_i[v])=i$.

\noindent
\textbf{Adversarial loss.}
We do not have the GT detailed shape for an arbitrary coarse shape $c$. Thus we resort to the discriminators to supervise shape generation. In addition, the geometric style of the generated shape should respect the designated styles $S(v)$ in the input voxels. Hence, we have the following adversarial loss to train the generators, which is a masked version of the LSGAN~\cite{mao2017least} loss,
\begin{align}
\mathcal{L}_{GAN} & = \mathbb{E}_{v} (
( D^j_*(G^j(c))[v] - 1 )^2 + \alpha \cdot ( D^j_{S(c[v])}(G^j(c))[v] - 1 )^2
),
\end{align}
where $D^j(\cdot)$ is the output voxel grid of the discriminator, and $\alpha$ is a parameter to control the influence of the style-specific discriminators.
Losses to train the discriminators can be found in the supplementary.

\noindent
\textbf{Adaptive $\bm{\alpha}$ weighting.}
Setting $\alpha$ to a larger value can make the generated shape more stylistic in the region near $v$ with respect to the style $S(c[v])$, yet it can make the region less plausible and less coherent with other regions, since the influence of the global discriminator has been tuned down. On the other hand, setting $\alpha$ to a small value lets the generator generate structurally coherent but style-less shapes.
Our observation is that the regions near the transition boundary where two parts of different styles meet are the most problematic, since the geometries in these regions are unlikely to be observed in our training examples with single styles.
Therefore, we develop a novel strategy to set $\alpha$ adaptively for each voxel: if a voxel is near a transition boundary (within 2 voxels), we will set a small $\alpha$ for that voxel, e.g., $\alpha_1 = 0.1$; otherwise, we set a larger $\alpha$, e.g., $\alpha_2 = 0.5$.

\noindent
\textbf{Structure-preserving losses.}
Finally, to make the generated shape respect the structures presented in the input coarse voxels, we propose two novel structure-preserving losses.
The downsampling loss ensures that if we downsample the generated shape at level $j+1$ from resolution $2^{j+1}$ to $2^{j}$, the result agrees with the generated shape at level $j$, or the input shape if $j=k$.
\begin{align}
\label{eqn:downsampling_loss}
    \mathcal{L}_{down} = ||\phi_{\downarrow}(G^{j+1}(c)) - \StopGradient(G^j(c))||_{2}^{2},
\end{align}
where $\phi_{\downarrow}$ is the max-pooling operator that downsamples the input by a factor of 2; $\StopGradient$ is the stop-gradient operator to prevent the generated shape in level $j+1$ from negatively affecting the generated shape in level $j$.
Similarly, we have an upsampling loss to ensure the upsampled result of level $j$ agrees with the shape at level $j+1$.
\begin{align}
\label{eqn:upsampling_loss}
    \mathcal{L}_{up} = ||G^{j+1}(c) - \StopGradient(\phi_{\uparrow}(G^{j}(c)))||_{2}^{2},
\end{align}
where $\phi_{\uparrow}$ upsamples the input by a factor of 2 via nearest neighbor.

The final loss is a sum of the above loss terms:
\begin{equation}
\label{eqn:final_loss}
    \mathcal{L} = \mathcal{L}_{recon} + \mathcal{L}_{GAN} + \gamma_1 \cdot \mathcal{L}_{down} + \gamma_2 \cdot \mathcal{L}_{up}.
\end{equation}
We set $\gamma_1=\gamma_2=10$ in our experiments.

\subsection{Data augmentation}
\label{sec:data_augmentation}
\begin{figure}[t]
\begin{picture}(244, 65)
  \put(0, 0){\includegraphics[width=1.00\linewidth]{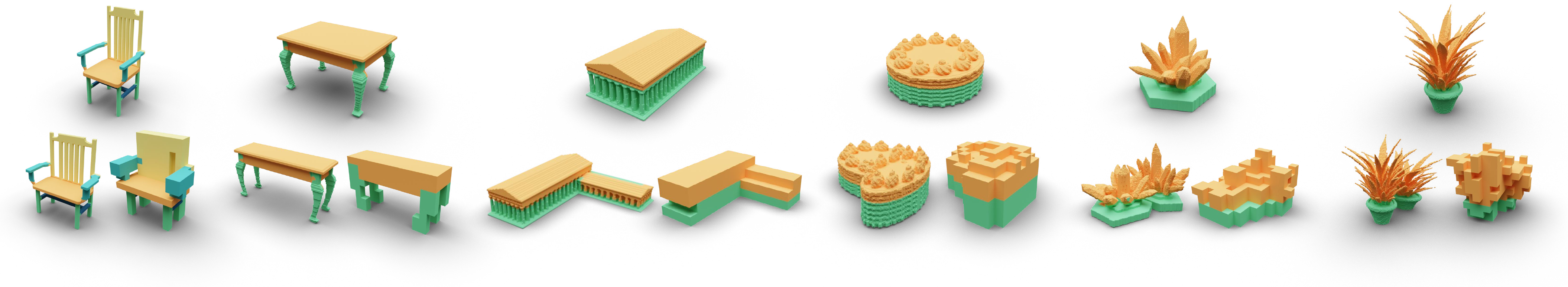}}
\end{picture}
\vspace{-3mm}
  \caption{Augmentation examples of different categories. For each category, we show the original style shape in the first row, the corresponding augmented style shape in the second row left, and downsampled as coarse shapes for training in the second row right.}
  \vspace{-3mm}
  \label{fig:augmentation_example}
\end{figure}

From few detailed and segmented style shapes, we synthesize an arbitrary number of coarse voxels with diverse structures and part segmentation for training, as shown in Figure~\ref{fig:augmentation_example}.
To create a coarse voxel, we choose a random style shape $s_i$ and randomly scale it in x, y, and z directions. For some categories such as plant, we further perform random rotation and combine multiple shapes to have more geometric variations. We downsample the augmented shapes into $2^k$ resolution as coarse shapes for training.
Note that all the style shapes are co-segmented, therefore we also obtain segmentation in the resulting coarse voxels. At this point, we can randomly assign different styles to different segmented parts. 

\subsection{Implementation details}
\label{sec:implementation_details}

Similar to DECOR-GAN~\cite{chen2021decor}, we apply a Gaussian filter with $\sigma=1$ on each style shape to convert its binary occupancy voxels into a smoother and more continuous scalar grid for the sake of better optimization. Training our model takes about 30 hours on a single NVIDIA 3090Ti GPU for $k=4$ and $K=8$. After training, generating a detailed shape only takes less than a second. We extract the mesh surfaces using Marching Cubes \cite{marchingcube1987}.
\section{Experiments}
\label{sec:exps}

In this section, we first evaluate our proposed method in single-category single-style shape detailization and compare with other detailization methods in Sec.~\ref{sec:single-category-single-style}. Next, we demonstrate that our method can generate novel 3D shapes with better localized style control in multi-category style mixing in Sec.~\ref{sec:multi-category-style-mixing}, and perform ablation study in Sec.~\ref{sec:ablation}. We also show several applications including interactive editing in Sec.~\ref{sec:application}.

\vspace{3mm}
\noindent
\textbf{Datasets.}
We conduct experiments on six shape categories: 16 chairs, 16 tables, and 5 plants from ShapeNet~\cite{chang2015shapenet}; and 5 buildings, 3 cakes, and 3 crystals from 3D Warehouse~\cite{3dwarehouse} under \href{https://creativecommons.org/licenses/by/4.0/}{CC-BY 4.0}. For each style shape, we obtain binary occupancy voxels and manually annotate the part labels for training. We segment each chair into five parts: back, seat, armrest, leg, and stretcher. Tables are labeled into two parts: tabletop and legs. A plant is labeled into pot and leaf. A building is labeled into the roof and main body. Cakes and crystals are segmented into bottom and top parts. All the training style shapes can be found in the supplementary. While increasing the number of style shapes is possible, we aim to demonstrate that our proposed method has \textit{robust} generalizability to coarse voxels with drastically different and complex structures, even when trained on a very limited number of style shapes.

\vspace{3mm}
\noindent
\textbf{Evaluation metrics.}
For quantitative evaluation, we follow DECOR-GAN~\cite{chen2021decor} and adopt the following metrics.
\textit{Strict-IOU} is to measure the Intersection over Union (IOU) between the downsampled output voxels and the input voxels, to evaluate how well the generated shape respects the structures in the input shape.
\textit{Loose-IOU} is a relaxed version of Strict-IOU to compute the proportion of occupied voxels in the input that are also occupied in the downsampled output.
\textit{LP-IOU} and \textit{LP-F-score} are to measure the percentage of local patches in the generated shape that are “similar” (according to IOU or F-score) to at least one local patch in the style shapes. Higher LP-IOU and LP-F-score indicate that the local details of the generated shapes are more similar to the local details of the style shapes, thus the generated shapes are deemed to be more locally plausible.
\textit{Cls-score} is to evaluate the overall plausibility of the generated shapes by training a classifier network to distinguish between the rendered images of the generated shapes and those of the real shapes and recording the mean classification accuracy.
More details can be found in the supplementary.

\subsection{Single-category detailization}
\label{sec:single-category-single-style}
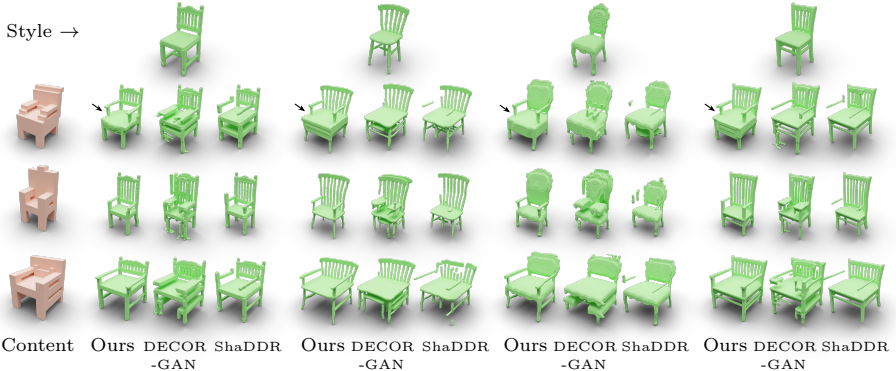
\begin{figure*}[t]
\begin{picture}(0, 135)
\centering
  \put(0, 0){\includegraphics[width=1.0\linewidth]{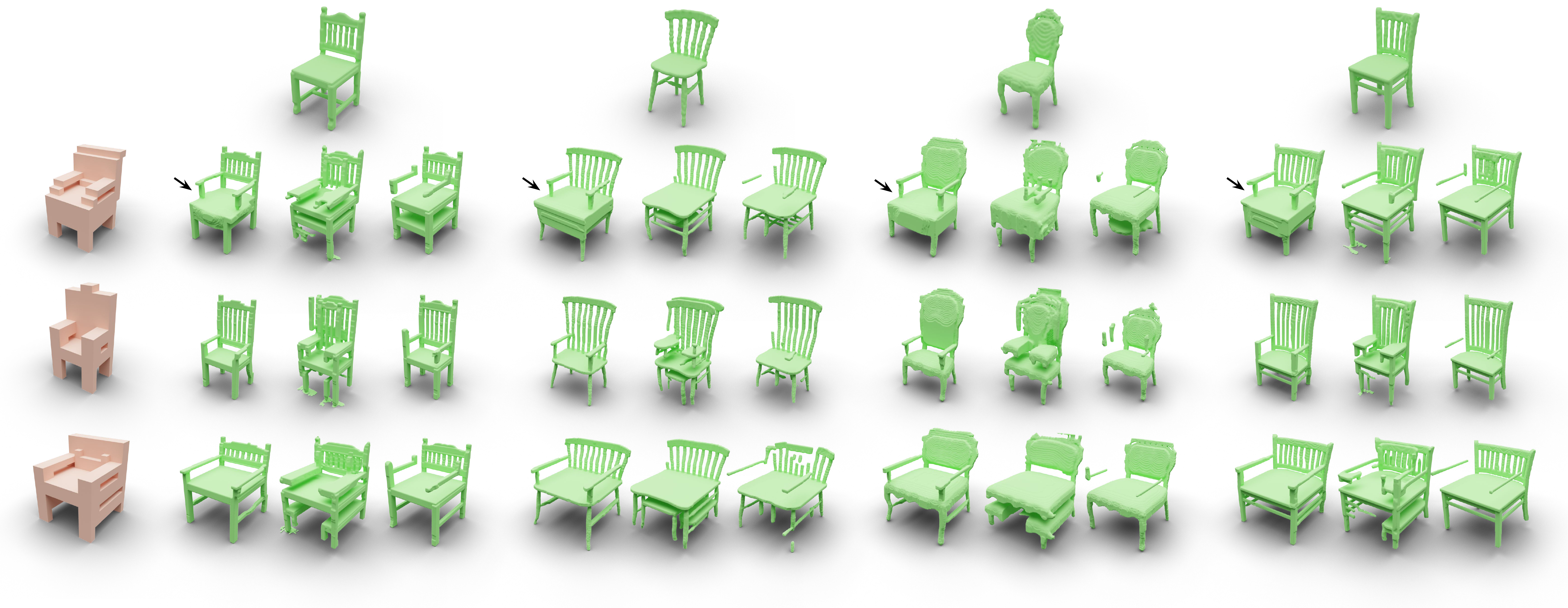}}
  \put(4, 3){\scriptsize Content}
  \put(6, 120){\scriptsize Style $\rightarrow$}
  
  \put(38, 3){\scriptsize Ours}
  \put(58, 3){\tiny DECOR}
  \put(61, -4){\tiny -GAN}
  \put(85, 3){\tiny ShaDDR}

  \put(118, 3){\scriptsize Ours}
  \put(137, 3){\tiny DECOR}
  \put(140, -4){\tiny -GAN}
  \put(164, 3){\tiny ShaDDR}

  \put(195, 3){\scriptsize Ours}
  \put(214, 3){\tiny DECOR}
  \put(217, -4){\tiny -GAN}
  \put(240, 3){\tiny ShaDDR}

  \put(271, 3){\scriptsize Ours}
  \put(290, 3){\tiny DECOR}
  \put(293, -4){\tiny -GAN}
  \put(316, 3){\tiny ShaDDR}
\end{picture}
  \caption{Single-category detailization on chair category. We show the input content shapes on the left and style shapes on top. Please zoom in to observe the details.}
  \vspace{-3mm}
  \label{fig:single_category_detailization}
\end{figure*}

We first qualitatively and quantitatively compare our method with DECOR-GAN~\cite{chen2021decor} and ShaDDR's geometry generator~\cite{chen2023shaddr} on single-category single-style shape detailization.
We use the data in DECOR-GAN and train individual models for different categories for fair comparisons.
We report the results on chair category due to page limit; other categories can be found in the supplementary.

We show quantitative comparison in Table~\ref{tab:single_category_single_style} and qualitative results in Figure~\ref{fig:single_category_detailization}.
DECOR-GAN and ShaDDR are likely to generate disconnected parts, especially in the joint regions, e.g., where armrests meet seats or backs.
Their results in Figure~\ref{fig:single_category_detailization} frequently show fragmented or disconnected parts.
In contrast, our method can produce significantly higher-quality upsampled geometry with better connectivity.
Moreover, our generated shapes better preserve the structures in the input voxels. An example is indicated by arrows in Figure~\ref{fig:single_category_detailization}, where the armrest of our generated chair closely follows the shape of its coarse content voxels, while the results of other methods fail to follow.
This is also reflected by our higher Strict- and Loose-IOU in Table~\ref{tab:single_category_single_style}.
Our method also generates better local details, as reflected by higher LP-IOU and LP-F-score.

\begin{table}
\vspace{-3mm}
\begin{minipage}{0.49\textwidth}
\centering
\caption{Quantitative results of single category detailization on the chair category.}
\label{tab:single_category_single_style}
\vspace{1.5mm}
\resizebox{1.0\columnwidth}{!}{
    \begin{tabular}{lccccc}
    \toprule
        & Strict- & Loose- & LP- & LP-F- & Cls- \\
        & IOU $\uparrow$ & IOU $\uparrow$ & IOU $\uparrow$ & score $\uparrow$ & score $\downarrow$ \\
    \midrule
       DECOR-GAN & 0.581 & 0.753 & 0.517 & 0.906 & 0.533 \\
       ShaDDR & 0.596 & 0.760 & 0.563 & 0.907 & 0.527  \\
       Ours (Pyramid full) & \textbf{0.748} & \textbf{0.908} & \textbf{0.591} & \textbf{0.914} & \textbf{0.506} \\
    \bottomrule
    \end{tabular}
    }
\end{minipage}
\begin{minipage}{0.49\textwidth}
\centering
\caption{Quantitative results of multi-category detailization on chair and table.}
\label{tab:multi_category_style_mixing}
\vspace{-1mm}
\resizebox{1.0\columnwidth}{!}{
    \begin{tabular}{lccccc}
    \toprule
        & Strict- & Loose- & LP- & LP-F- & Cls- \\
        & IOU $\uparrow$ & IOU $\uparrow$ & IOU $\uparrow$ & score $\uparrow$ & score $\downarrow$ \\
    \midrule
       DECOR-GAN$^{*}$ & 0.609 & 0.839 & 0.509 & 0.961 & 0.567 \\
       ShaDDR$^{*}$ & 0.611 & 0.854 & 0.496 & 0.933 & 0.549 \\
       Ours \textit{w/o} part labels & 0.747 & 0.900 & 0.508 & 0.969 & 0.533 \\
       Ours \textit{w} part labels & \textbf{0.750} & \textbf{0.906} & \textbf{0.513} & \textbf{0.977} & \textbf{0.518} \\
    \bottomrule
    \end{tabular}
    }
\end{minipage}
\vspace{-8.5mm}
\end{table}

\subsection{Multi-category style mixing}
\label{sec:multi-category-style-mixing}
\begin{figure*}[t]
\begin{picture}(0, 160)
\centering
  \put(0, 0){\includegraphics[width=1.0\linewidth]{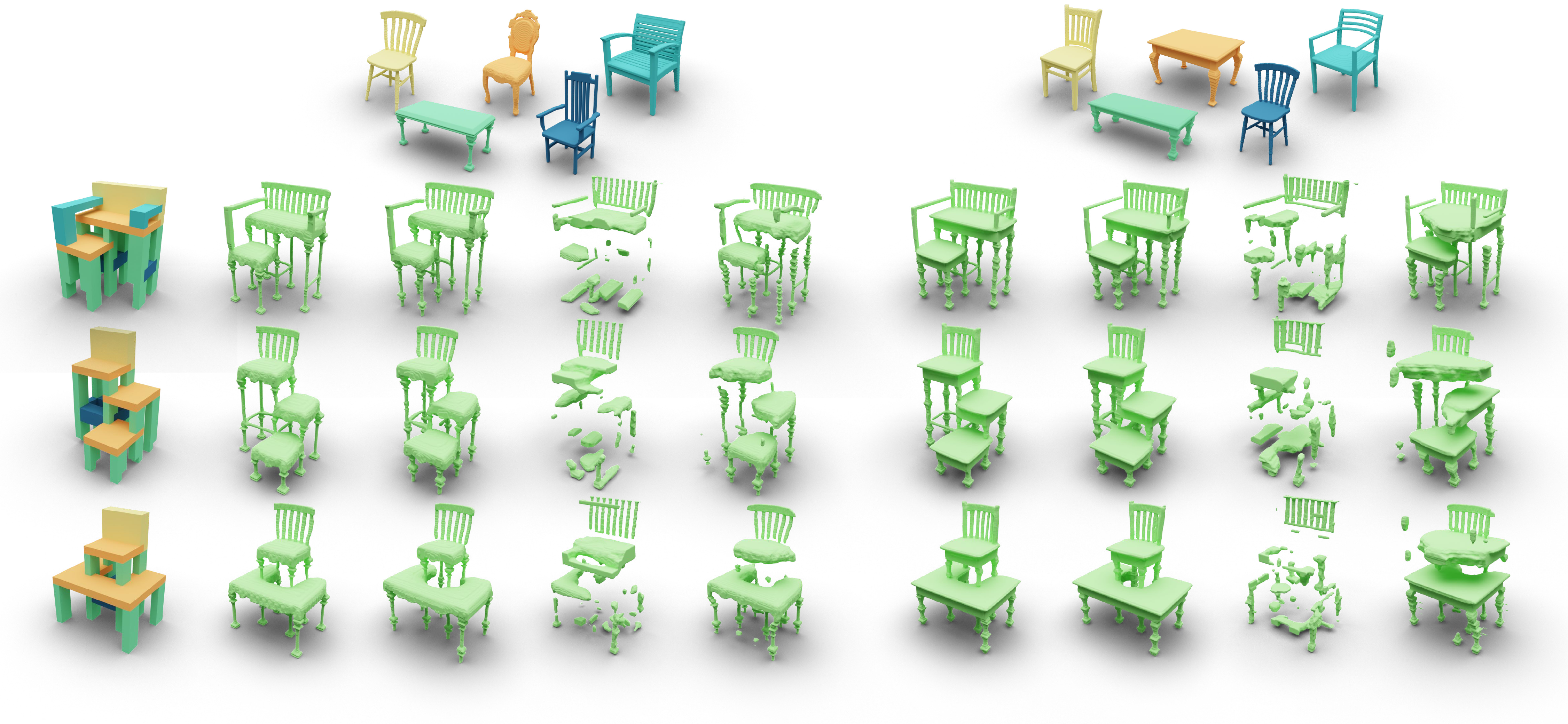}}
  \put(10, 4){\scriptsize Content}
  \put(13, 140){\scriptsize Styles $\rightarrow$}
  
  \put(50, 4){\tiny Ours \textit{w}}
  \put(43, -3){\tiny part labels}
  \put(84, 4){\tiny Ours \textit{w/o}}
  \put(83, -3){\tiny part labels}
  \put(122, 4){\tiny DECOR}
  \put(125, -3){\tiny -GAN$^{*}$}
  \put(155, 4){\tiny ShaDDR$^{*}$}

  \put(203, 4){\tiny Ours \textit{w}}
  \put(198, -3){\tiny part labels}
  \put(237, 4){\tiny Ours \textit{w/o}}
  \put(236, -3){\tiny part labels}
  \put(276, 4){\tiny DECOR}
  \put(279, -3){\tiny -GAN$^{*}$}
  \put(310, 4){\tiny ShaDDR$^{*}$}


\end{picture}
  \caption{Multi-category style mixing results on chair and table categories. We show the input coarse voxels with style labels on the left. The corresponding style shapes for the colored style labels are shown on top. Please zoom in to observe the details.}
  \vspace{-3mm}
  \label{fig:multi_category_style_mixing_chair_table}
\end{figure*}
\begin{figure*}[t]
\begin{picture}(0, 160)
\centering
  \put(0, 0){\includegraphics[width=1.0\linewidth]{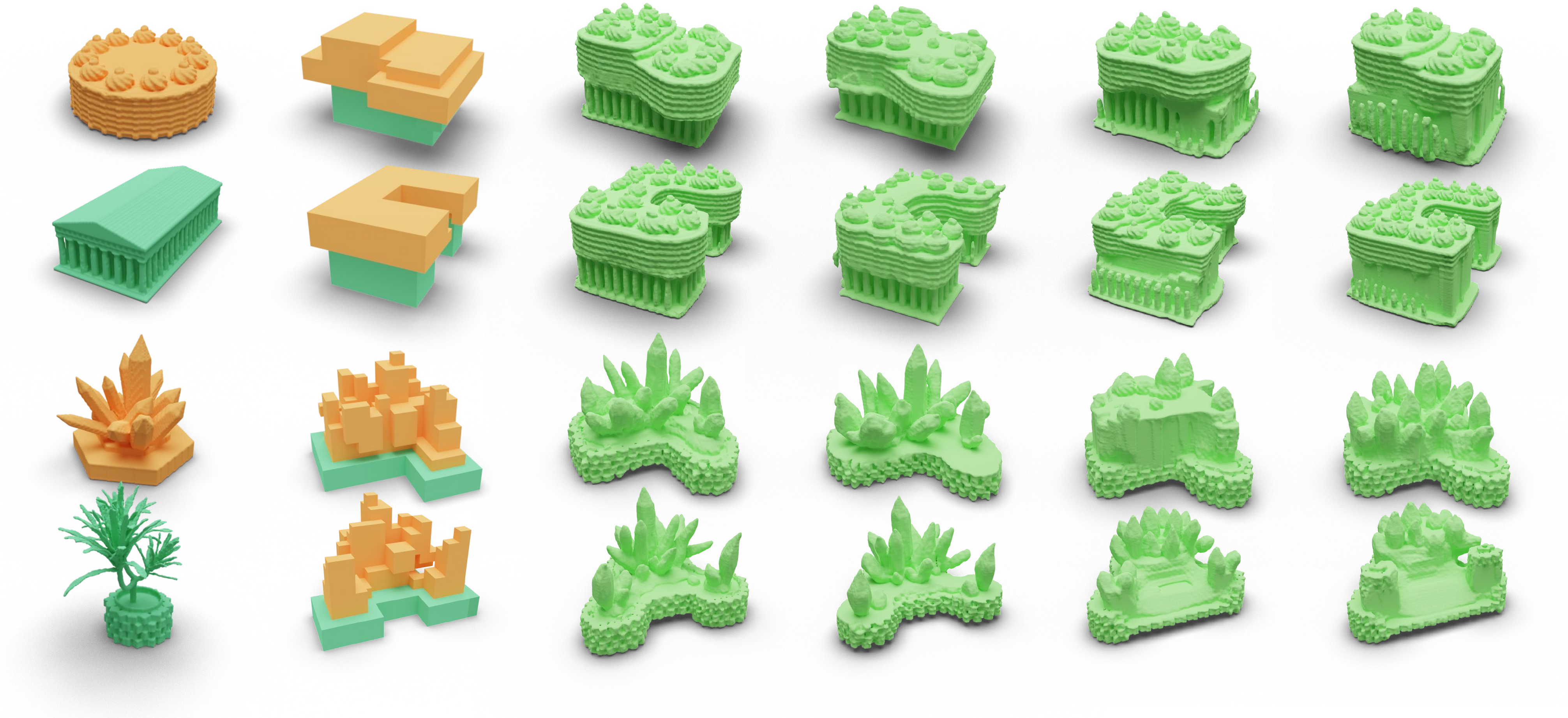}}
  \put(20, 4){\scriptsize Styles}
  \put(73, 4){\scriptsize Content}
  \put(135, 4){\scriptsize Ours \textit{w}}
  \put(130, -3){\scriptsize part labels}
  \put(187, 4){\scriptsize Ours \textit{w/o}}
  \put(185, -3){\scriptsize part labels}
  \put(237, 4){\scriptsize DECOR-GAN$^{*}$}
  \put(300, 4){\scriptsize ShaDDR$^{*}$}

  \put(0, 120){\scriptsize (a)}
  \put(0, 45){\scriptsize (b)}
\end{picture}
  \caption{Multi-category style mixing results on plant, building, cake, and crystal categories. For each group, e.g., (a), we show two style shapes in the first column, coarse input shapes with style labels in the second column, and results in the remaining columns. Please zoom in to observe the local geometric details.}
  \vspace{-3mm}
  \label{fig:multi_category_style_mixing_pbcc}
\end{figure*}


We train one single model with $k=4$ and $K=8$ for the chair and table categories and another with $k=4$ and $K=8$ for the plant, building, cake and crystal categories. 
Since the original DECOR-GAN and ShaDDR are not able to perform style mixing during geometry detailization, we modify the training procedure of DECOR-GAN and ShaDDR by assigning different styles to different regions of the input coarse voxels for fair comparisons. We denote these two baselines as DECOR-GAN$^{*}$ and ShaDDR$^{*}$.

We show the style mixing results in Figure~\ref{fig:multi_category_style_mixing_chair_table} and \ref{fig:multi_category_style_mixing_pbcc}. DECOR-GAN$^{*}$ and ShaDDR$^{*}$ fail to produce coherent and connected structures for out-of-distribution content shapes with different style combinations. Our method not only demonstrates improved generalization to novel coarse content shapes, but also produces better geometric details with smooth style transition.
Our method also outperforms other methods quantitatively, as shown in Table~\ref{tab:multi_category_style_mixing}.
More qualitative results can be found in the supplementary material.

\begin{figure}[ht]
\begin{minipage}{.49\textwidth}
\begin{picture}(0, 95)
  \put(0, 0){\includegraphics[width=\textwidth]{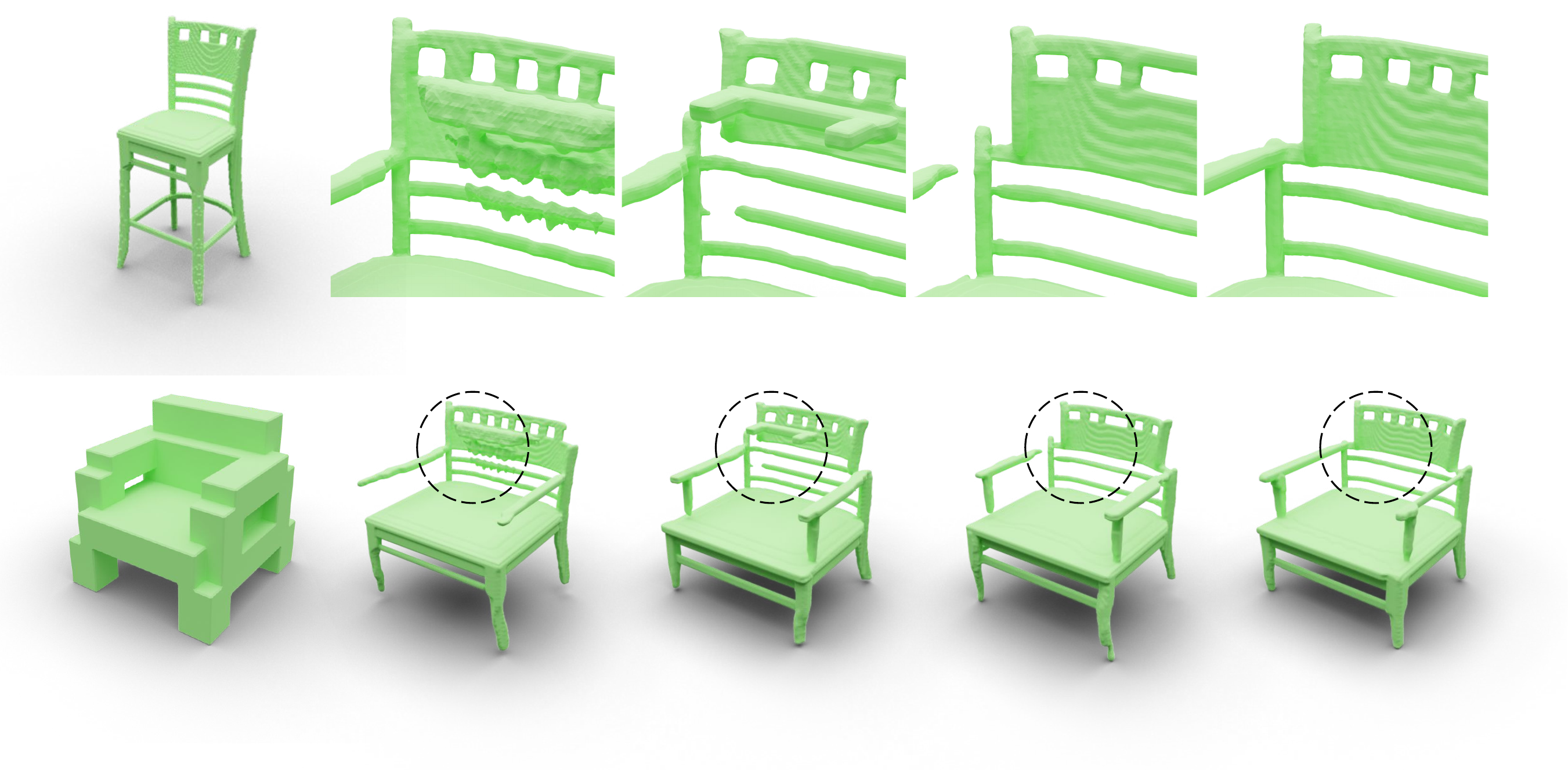}}
  \put(6, 5){\scriptsize Content}
  \put(12, 45){\scriptsize Style}

  \put(43, 5){\scriptsize vanilla}
  \put(72, 5){\scriptsize \textit{w} \tiny $\mathcal{L}_{up}$}
  \put(102, 5){\scriptsize \textit{w} \tiny $\mathcal{L}_{down}$}
  \put(143, 5){\scriptsize full}

  \put(48, 45){\scriptsize (a)}
  \put(80, 45){\scriptsize (b)}
  \put(112, 45){\scriptsize (c)}
  \put(143, 45){\scriptsize (d)}
\end{picture}
  \vspace{-3mm}
  \caption{Qualitative ablation study of the proposed structure-preserving losses in the single-category setting.}
  \label{fig:single_category_qualitative_ablation}
\end{minipage}
\begin{minipage}{.49\textwidth}
    \captionof{table}{Quantitative ablation study of the proposed structure-preserving losses in the single-category setting.}
    \vspace{4mm}
    \resizebox{1.0\columnwidth}{!}{
    \begin{tabular}{lccccc}
    \toprule
        & Strict- & Loose- & LP- & LP-F- & Cls- \\
        & IOU $\uparrow$ & IOU $\uparrow$ & IOU $\uparrow$ & score $\uparrow$ & score $\downarrow$ \\
    \midrule
    \vspace{1mm}
       (a). Pyramid vanilla & 0.578 & 0.819 & 0.508 & 0.868 & 0.593  \\
       \vspace{1mm}
       (b). Pyramid \textit{w} $\mathcal{L}_{up}$ & 0.705 & 0.873 & 0.547 & 0.883 & 0.550  \\
       \vspace{1mm}
       (c). Pyramid \textit{w} $\mathcal{L}_{down}$ & 0.730 & 0.889 & 0.564 & 0.892 & 0.558  \\
       (d). Pyramid full & \textbf{0.748} & \textbf{0.908} & \textbf{0.591} & \textbf{0.914} & \textbf{0.506}  \\
    \bottomrule
    \end{tabular}
    \label{tab:single_category_quantitative_ablation}
    }
\end{minipage}
\vspace{-3mm}
\end{figure}

\subsection{Ablation study}
\label{sec:ablation}
In this section, we validate the effectiveness of our pyramid GAN structure and structure-preserving losses. In Table~\ref{tab:single_category_quantitative_ablation} and Figure~\ref{fig:single_category_qualitative_ablation}, we compare several variations of our proposed method in single-category single-style setting.
(1) \textit{Pyramid vanilla}, in which we only use a pyramid GAN structure without structure-preserving losses, i.e. $\gamma_1 = \gamma_2 = 0$ in Equation \ref{eqn:final_loss}. In this setting, we adopt DECOR-GAN's design for preserving the content structure, which are non-differentiable masks applied to the generated voxels to cut off all the voxels outside the valid region defined by the coarse input voxels. We observe connectivity issues as well as floating pieces in the generated shape, as shown in Figure~\ref{fig:single_category_qualitative_ablation} (a).
(2) \textit{Pyramid w} $\mathcal{L}_{up}$, in which we remove the masks and add the upsampling loss to Pyramid vanilla, i.e., $\gamma_1=0$, $\gamma_2=10$. The upsampling loss can better preserve the overall structure, while the floating pieces remain, as shown in Figure~\ref{fig:single_category_qualitative_ablation} (b).
(3) \textit{Pyramid w} $\mathcal{L}_{down}$, in which we remove the masks and add the downsampling loss to Pyramid vanilla, i.e., $\gamma_1=10$, $\gamma_2=0$. The downsampling loss can also help preserve the overall structure, and it effectively eliminates the floating pieces. Yet it tends to miss thin structures, as shown in Figure~\ref{fig:single_category_qualitative_ablation} (c).
(4) Our proposed method, \textit{Pyramid full}, where $\gamma_1=\gamma_2=10.0$. In this setting, the generated shapes exhibit better connectivity and better adherence to the global structure, as shown in Figure~\ref{fig:single_category_qualitative_ablation} (d).
The quantitative results in Table~\ref{tab:single_category_quantitative_ablation} also show that our full model has the best performance.

\begin{table}[t]
    \begin{center}
    \caption{Quantitative ablation studies of Pyramid structures (P), structure-preserving losses ($\mathcal{L}_{down}$ and $\mathcal{L}_{up}$) and adaptive $\alpha$ weighting (adp-$\alpha$) in the multi-category chair and table style mixing setting.}
    \label{tab:detailed_multi_category_ablation}
    \vspace{-3mm}
    \resizebox{1.0\columnwidth}{!}{
    \begin{tabular}{lccccc}
    \toprule
        & \small Strict-IOU $\uparrow$ & \small Loose-IOU $\uparrow$ & \small LP-IOU $\uparrow$ & \small LP-F-score $\uparrow$ & \small Cls-score $\downarrow$ \\
    \midrule
       \textit{(a)}. P \smallxmark \;(DECOR-GAN$^{*}$) & 0.621 & 0.849 & 0.261 & 0.939 & 0.638 \\
       \textit{(b)}. P \smallcheck, $\mathcal{L}_{down}$ \smallxmark, $\mathcal{L}_{up}$\smallxmark, adp-$\alpha$ \smallxmark & 0.661 & 0.880 & 0.260 & 0.941 & 0.645 \\
       
       \textit{(c)}. P \smallcheck, $\mathcal{L}_{down}$ \smallcheck, $\mathcal{L}_{up}$\smallxmark, adp-$\alpha$ \smallxmark & 0.713 & 0.894 & 0.274 & 0.949 & 0.586 \\
       \textit{(d)}. P \smallcheck, $\mathcal{L}_{down}$ \smallxmark, $\mathcal{L}_{up}$\smallcheck, adp-$\alpha$ \smallxmark & 0.705 & 0.889 & 0.271 & 0.952 & 0.592 \\
       \textit{(e)}. P \smallcheck, $\mathcal{L}_{down}$ \smallxmark, $\mathcal{L}_{up}$\smallxmark, adp-$\alpha$ \smallcheck & 0.688 & 0.882 & 0.263 & 0.955 & 0.568 \\
       
       \textit{(f)}. P \smallcheck, $\mathcal{L}_{down}$ \smallcheck, $\mathcal{L}_{up}$\smallcheck, adp-$\alpha$ \smallxmark & 0.753 & 0.905 & 0.281 & 0.963 & 0.533 \\
       \textit{(g)}. P \smallcheck, $\mathcal{L}_{down}$ \smallcheck, $\mathcal{L}_{up}$\smallxmark, adp-$\alpha$ \smallcheck & 0.721 & 0.904 & 0.279 & 0.960 & 0.529 \\
       \textit{(h)}. P \smallcheck, $\mathcal{L}_{down}$ \smallxmark, $\mathcal{L}_{up}$\smallcheck, adp-$\alpha$ \smallcheck & 0.724 & 0.898 & 0.277 & 0.952 & 0.547 \\

       \textit{(\,i)}. P \smallcheck, $\mathcal{L}_{down}$ \smallcheck, $\mathcal{L}_{up}$\smallcheck, adp-$\alpha$ \smallcheck & \textbf{0.761} & \textbf{0.913} & \textbf{0.282} & \textbf{0.968} & \textbf{0.527} \\
    \bottomrule
    \end{tabular}
    }
    \end{center}
    \vspace{-6mm}
\end{table}

\begin{figure*}[t]
\begin{picture}(0, 95)
\centering
  \put(0, 0){\includegraphics[width=1.0\linewidth]{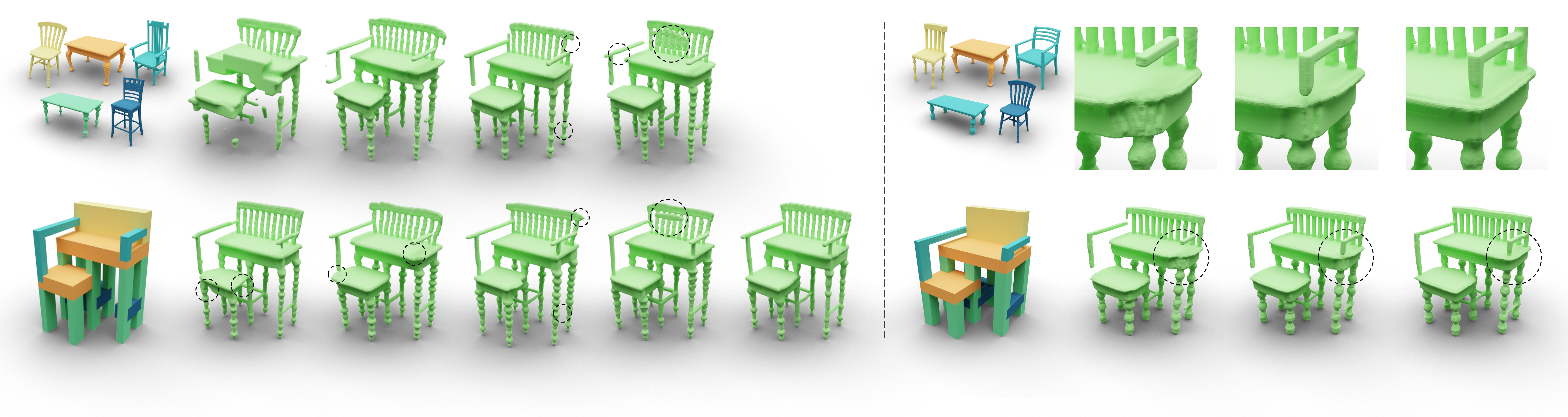}}
  \put(6, 6){\scriptsize Content}
  \put(11, 51){\scriptsize Styles}
  
  \put(50, 51){\scriptsize (a)}
  \put(80, 51){\scriptsize (b)}
  \put(110, 51){\scriptsize (c)}
  \put(140, 51){\scriptsize (d)}
  
  \put(50, 6){\scriptsize (e)}
  \put(80, 6){\scriptsize (f)}
  \put(110, 6){\scriptsize (g)}
  \put(140, 6){\scriptsize (h)}
  \put(170, 6){\scriptsize (i)}

  \put(202, 6){\scriptsize Content}
  \put(205, 51){\scriptsize Styles}

  \put(240, 10){\scriptsize $\alpha_{1}=0.5$}
  \put(240, 2){\scriptsize $\alpha_{2}=0.5$}

  \put(275, 10){\scriptsize $\alpha_{1}=0.3$}
  \put(275, 2){\scriptsize $\alpha_{2}=0.5$}

  \put(312, 10){\scriptsize $\alpha_{1}=0.1$}
  \put(312, 2){\scriptsize $\alpha_{2}=0.5$}

\end{picture}
\vspace{-3mm}
  \caption{\textbf{(Left)} Qualitative ablation studies of Pyramid structures (P), structure-preserving losses ($\mathcal{L}_{down}$ and $\mathcal{L}_{up}$) and adaptive $\alpha$ weighting (adp-$\alpha$) in the multi-category chair and table style mixing setting. The configurations of different models used in (a-i) can be found in Table~\ref{tab:detailed_multi_category_ablation}. Zoom in to observe the details. \textbf{(Right)} Qualitative ablation study of different adaptive $\alpha$ values. The input coarse voxel and style shapes are shown on the left. The zoom-ins are shown on the top. $\alpha_{1}$ is set for voxels near the transition boundary and $\alpha_{2}$ for the rest.}
  \label{fig:multi_category_ablation_alpha_ablation}
  \vspace{-4mm}
\end{figure*}

We also perform a more thorough ablation study of each component in the multi-category style mixing setting, as shown in Table~\ref{tab:detailed_multi_category_ablation} and Figure~\ref{fig:multi_category_ablation_alpha_ablation} (left). Note that adaptive $\alpha$ is only used in the multi-category style mixing setting. By leveraging Pyramid architecture ((a) vs. (b)), the overall structure of the output is significantly improved. Both $\mathcal{L}_{down}$ and $\mathcal{L}_{up}$ further help \textit{refine} the overall structure where $\mathcal{L}_{down}$ can effectively eliminate the floating pieces and $\mathcal{L}_{up}$ can improve thin structures ((c) vs. (d) and (g) vs. (h)). This conclusion is also consistent with the ablation study in the single-category detailization setting. The adaptive $\alpha$ weighting can effectively improve the boundary \textit{transition} where two different styles meet, e.g. the armrest is well-connected to the seat and the stretcher is well-connected to the leg ((c) vs. (g), (d) vs. (h) and (f) vs. (i)).

In addition, we perform an ablation study on the adaptive $\alpha$ weighting scheme described in Sec.~\ref{sec:loss_functions} with different $\alpha$ values. Figure~\ref{fig:multi_category_ablation_alpha_ablation} (right) shows qualitative results of setting the parameter $\alpha_{1}$ to $0.1$, $0.3$ and $0.5$ for voxels near the transition boundary and $\alpha_{2}$ to $0.5$ for the rest regions. Setting $\alpha_{1}=\alpha_{2}=0.5$ can be considered equivalent to not adaptively adjusting the $\alpha$. By using a smaller $\alpha_1$, the region where two different styles meet has a smoother style transition, e.g. the armrest is well-connected to the seat in the last column of Figure~\ref{fig:multi_category_ablation_alpha_ablation} (right).

\begin{figure*}[t]
\begin{picture}(0, 120)
  \put(0, 0){\includegraphics[width=0.99\linewidth]{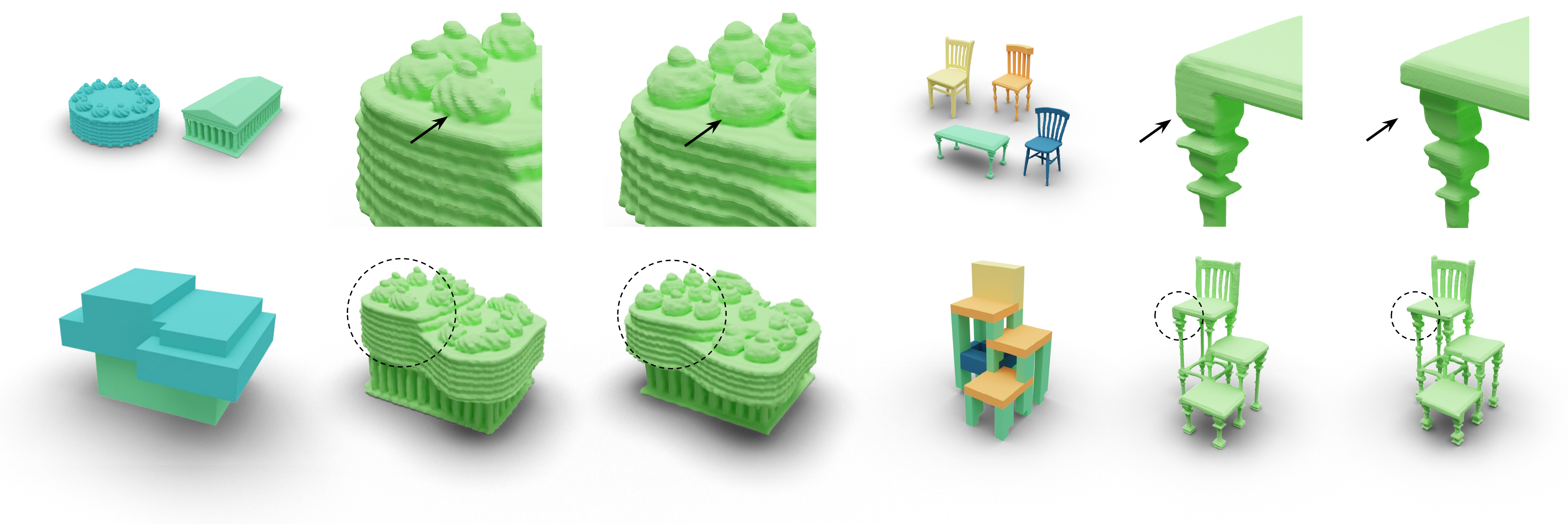}}
  \put(25, 5){\scriptsize Content}
  \put(27, 67){\scriptsize Styles}

  \put(90, 9){\scriptsize \textit{with}}
  \put(80, 0){\scriptsize part labels}
  \put(146, 9){\scriptsize \textit{without}}
  \put(142, 0){\scriptsize part labels}

  \put(205, 5){\scriptsize Content}
  \put(207, 67){\scriptsize Styles}

  \put(260, 9){\scriptsize \textit{with}}
  \put(250, 0){\scriptsize part labels}
  \put(305, 9){\scriptsize \textit{without}}
  \put(301, 0){\scriptsize part labels}
\end{picture}
  \caption{Ablation study of input \textit{with} vs.~\textit{without} part labels on building-cake and chair-table style mixing. By leveraging part labels as additional input, the network can generate more details with more natural connections between different regions.}
  \vspace{-4mm}
  \label{fig:part_labels_ablation}
\end{figure*}

We further stress test our method by removing part labels from the input to the network. That is, the users need not specify which part of each exemplar should supply details: the network needs to automatically decide this based on the geometry of the selected region on the coarse shape. Therefore, no co-segmentation is needed in this setting, and we only use per-shape segmentation to assign styles to each local region of the coarse input shape during training. 

Figure~\ref{fig:part_labels_ablation} visually compares inputs \textit{with} vs.~\textit{without} part labels. Even without part labels, our method can generate reasonable results with only slightly worse local details, which may be attributed to the network leveraging certain capabilities to identify selected regions. This is also reflected by slightly lower LP-IOU and LP-F-score compared to input \textit{with} part labels in Table~\ref{tab:multi_category_style_mixing}.

\subsection{Application}
\label{sec:application}

Our method can be applied to detailize shapes from various sources, as shown in Figure~\ref{fig:application}.
(a) We can detailize coarse shapes that are easily obtainable by extruding 2D profiles, such as fonts.
(b) We develope an interface for users to model coarse voxels and assign style labels interactively. Please see our supplementary video for a real-time, interactive demo.
(c) We ran an off-the-shelf text-to-3D model to obtain a ``C-shaped cake''. Then we remove the bottom plane of the generated shape and apply our method on its downsampled voxels to obtain a detailed cake.
(d) We can also detailize shapes that are created using simple primitives such as two boxes.
For each shape, we apply the styles of two cakes at different regions, which may not correspond to semantic parts.
Note that we offer the first method for interactive \textit{controllable} and \textit{localized} detail generation, unlike ShaDDR~\cite{chen2023shaddr} which only offers global style control. This may enable creative modeling, such as the building and plant mixing in Figure~\ref{fig:limitation} (a).
\section{Conclusion, discussion, and future work}
\label{sec:conclusions}

\begin{wrapfigure}{r}{0.35\textwidth}
\vspace{-1mm}
\begin{picture}(0, 60)
\centering
  \put(0, 0){\includegraphics[width=1.0\linewidth]{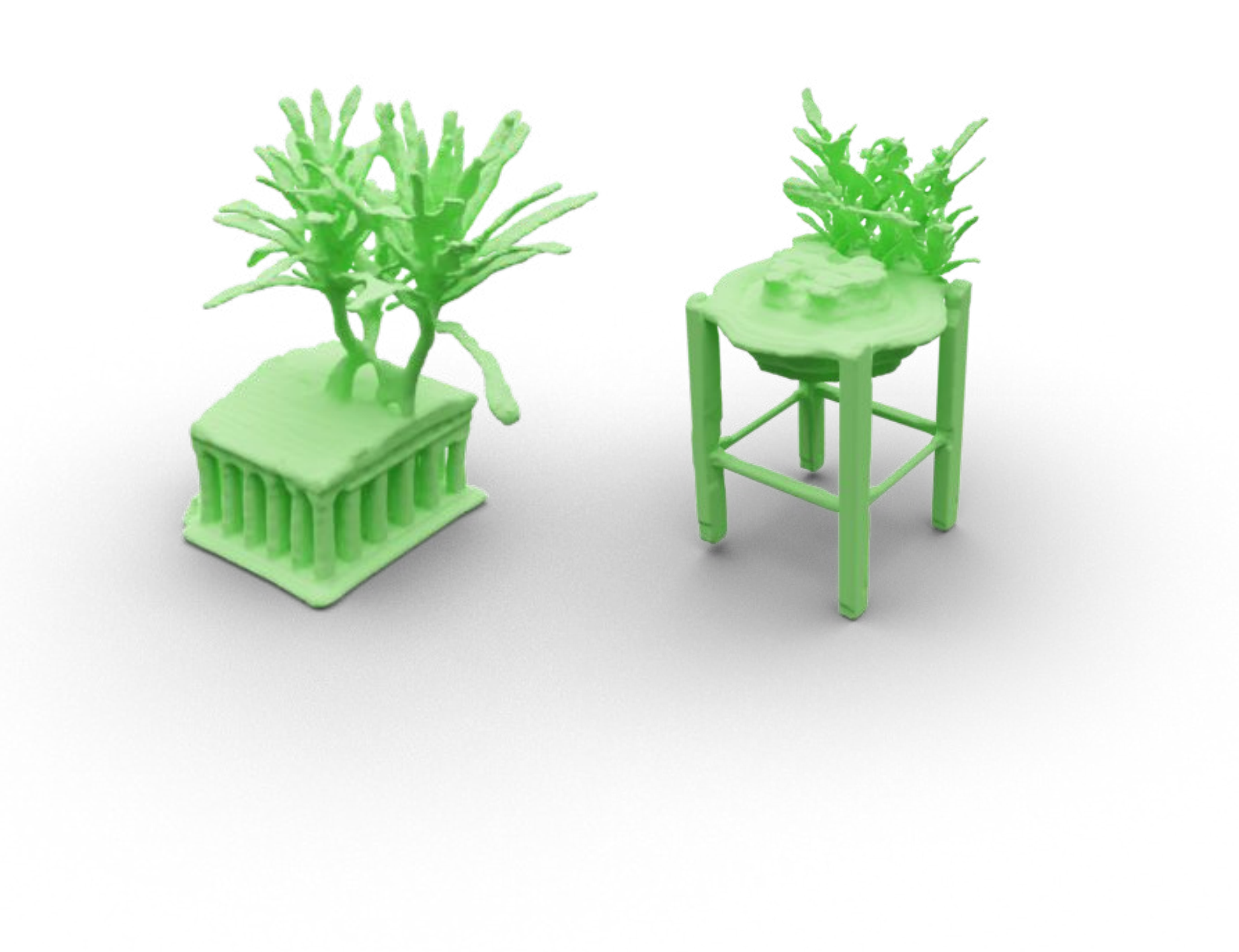}}
  \put(10, 18){\scriptsize (a) Creative}
  \put(23, 9){\scriptsize modeling.}

  \put(66, 18){\scriptsize (b) A failure}
  \put(80, 9){\scriptsize case.}
\end{picture}
\vspace{-4mm}
\caption{Examples of creative modeling and a failure case.}
\vspace{-7mm}
\label{fig:limitation}
\end{wrapfigure}

We present the first exemplar-based generative model for 3D detailization \qm{\textit{which}} offers local control, interactivity, and generalizability to out-of-distribution coarse structures. 
Our novel structure-preserving losses, along with the global discriminator and spatially adaptive style adjustment, lead to clear improvements over current detailization methods and enable coherent style transition even when mixing diverse exemplars.

To generate better local geometric details, we currently assume that a meaningful co-segmentation is available for all style shapes. This may limit the detailization to only coarse-level structures and also prevent style transfer which breaks the semantic barrier.
Our method adopts the occupancy voxel representation for 3D shapes and relies on 3D CNNs to perform upsampling, which can limit the resolution of the final generated shapes.
For example, our maximum output resolution is $256^3$, which may not be sufficient to represent finer geometric details, such as the tips of the crystals in Figure~\ref{fig:multi_category_style_mixing_pbcc} (b).
%
In our experiments, we found mixing styles for significantly different geometries, e.g., the failure case of mixing chair and plant in Figure~\ref{fig:limitation} (b), can often lead to undesirable artifacts. \qm{This is due to the significant differences in the \textit{local} structure of the coarse shape and the style shapes. For example, the chair back cannot be detailed into a plant because such a shape does not exist in either the content or the style shapes}. Mixing these styles requires a deeper understanding of semantics and aesthetics.

As for future work, we would like to transfer non-homogeneous shape details onto non-homogenous coarse structures.
The use of diffusion models for voxel upsampling and the integration with large language and vision-language models for geometry and texture detailization are both worth exploring.


\section*{Acknowledgments}

\qm{We thank the anonymous reviewers for their valuable comments. This work was done during the first author's internship at Adobe Research, and it is supported in part by an NSERC grant (No. 611370) and a gift fund from Adobe Research.}

%
%
\bibliographystyle{splncs04}
\bibliography{main}

\clearpage
\begin{center}
\textbf{\Large\ApproachName{}: 3D Detailization by Controllable, Localized, and Learned Geometry Enhancement} \\
\Large{(Supplementary Material)} \\
\end{center}

\setcounter{page}{1}

\appendix

\section{Data and code}

\begin{figure*}
\begin{picture}(0, 450)
\centering
  \put(0, 0){\includegraphics[width=0.99\linewidth]{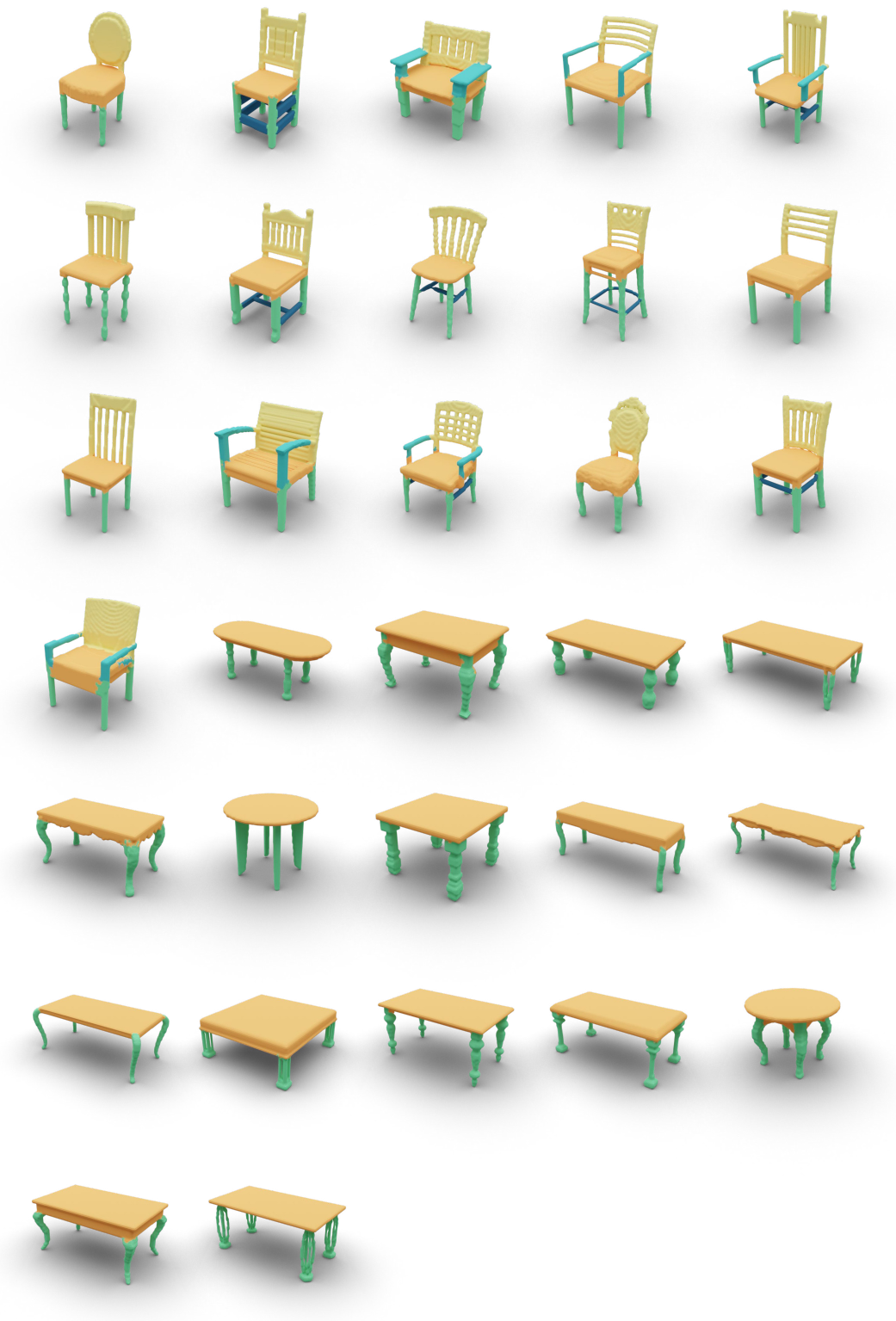}}
\end{picture}
  \caption{The segmented training style shapes of chair and table categories. We segment the chair into five parts: back, seat, leg, armrest and stretcher and the table into two parts: tabletop and leg.}
  \label{fig:chair_table_style_shapes}
\end{figure*}

We show the style shapes for training chair and table style mixing in Figure \ref{fig:chair_table_style_shapes}, and plant, building, cake and crystal in Figure \ref{fig:plant_building_cake_crystal_style_shapes}. We also provide the ready-to-use data and code in the supplementary. Code will be released upon acceptance.

\section{Loss function}

We follow the notations defined in the main paper and provide the discriminator loss.

\noindent
\textbf{Discriminator loss.}
For any style shape $s$ and any coarse shape $c$ with designated styles $S(v)$, the discriminator loss is the sum of the global discriminator $D_{*}^{j}$’s loss and the style-specific discriminators $D_{i}^{j}$’s loss:
\begin{align}
    L_{D^{j}} = L_{D_{*}^{j}} + L_{D_{style}^{j}}
\end{align}
where
\begin{align}
    L_{D_{*}^{j}} &= \mathbb{E}_{v} ( ( D_{*}^{j}(s^{j})[v] - 1 )^{2} +(D_{*}^{j}(G^{j}(c))[v])^{2} )
\end{align}
\begin{align}
    L_{D_{style}^{j}} &= \mathbb{E}_{v} ( (D_{S(s[v])}^{j}(s^{j})[v] - 1)^{2} \nonumber \\ &+ (D_{S(c[v])}^{j}(G^{j}(c))[v])^{2} )
\end{align}

\section{Evaluation metrics}

We use the following metrics from DECOR-GAN~\cite{chen2021decor} to quantitatively evaluate the quality of the generated shapes on both single-category detailization and multi-category style mixing tasks.
%

\noindent
\textbf{Strict-IOU and Loose-IOU.}
Strict-IOU is to measure the Intersection over Union (IOU) between the downsampled output voxels and the input voxels, to evaluate how well the generated shape respects the structures in the input shape.
Loose-IOU is a relaxed version of Strict-IOU to compute the proportion of occupied voxels in the input that are also occupied in the downsampled output.
%

\noindent
\textbf{LP-IOU and LP-F-score.}
LP-IOU and LP-F-score are to measure the percentage of local patches in the generated shape that are “similar” (according to IOU or F-score) to at least one local patch in the style shapes. Higher LP-IOU and LP-F-score indicate that the local details of the generated shapes are more similar to the local details of the style shapes, thus the generated shapes are deemed to be more locally plausible. We mark the two patches as “similar” if the IOU (F-score) is above 0.95. To reduce the computational complexity, we sample $12^3$ patches in a voxel model, a size slightly smaller than the receptive field of the discriminator. Additionally, We only sample surface patches with at least one occupied voxel and one unoccupied voxel in their central $2^3$ areas to avoid sampling featureless patches located far inside or outside the shape. We sample 1000 patches from each testing shape and compare them with all potential patches in the detailed shapes.
%

\noindent
\textbf{Cls-score.}
Cls-score is to evaluate the overall plausibility of the generated shapes by training a classifier network to distinguish between the rendered images of the generated shapes and those of the real shapes and recording the mean classification accuracy.
We train a ResNet using high-resolution voxels (from which content shapes are downsampled) as real samples and our generated shapes as fake samples. For each sample, we randomly render 24 images with a resolution of $256^2$. The images are randomly cropped to
10 small patches with a resolution of $64^2$ for training.

\noindent
\textbf{Evaluation details.} For IOU and LP, we evaluate 320 generated shapes (20 contents × 16 styles) since they are computationally expensive. For Cls-score, we evaluate 1600 generated shapes (100 contents × 16 styles). For multi-category style mixing, we generate 16 style sets, each containing 5 random style combinations.

\section{More results on single category detailization}

We show qualitative and quantitative results on the building category in Figure \ref{fig:single_category_building_plant} and Table \ref{tab:single_category_single_style_building}, and the plant category in Figure \ref{fig:single_category_building_plant} and Table \ref{tab:single_category_single_style_plant}.

\begin{table}
\begin{minipage}{0.49\textwidth}
\centering
\caption{Quantitative of single-category detailization on the building category.}
\label{tab:single_category_single_style_building}
\resizebox{1.0\columnwidth}{!}{
    \begin{tabular}{lccccc}
    \toprule
        & Strict- & Loose- & LP- & LP-F- & Cls- \\
        & IOU $\uparrow$ & IOU $\uparrow$ & IOU $\uparrow$ & score $\uparrow$ & score $\downarrow$ \\
    \midrule
       DECOR-GAN & 0.693 & 0.973 & 0.429 & \textbf{0.662} &  0.598 \\
       ShaDDR & 0.601 & 0.957 & 0.425 & 0.633 & 0.633  \\
       Ours & \textbf{0.732} & \textbf{0.987} & \textbf{0.442} & 0.648 & \textbf{0.592}  \\
    \bottomrule
    \end{tabular}
    }
\end{minipage}
\begin{minipage}{0.49\textwidth}
\centering
\caption{Quantitative of single-category detailization on the plant category.}
\label{tab:single_category_single_style_plant}
\resizebox{1.0\columnwidth}{!}{
    \begin{tabular}{lccccc}
    \toprule
        & Strict- & Loose- & LP- & LP-F- & Cls- \\
        & IOU $\uparrow$ & IOU $\uparrow$ & IOU $\uparrow$ & score $\uparrow$ & score $\downarrow$ \\
    \midrule
       DECOR-GAN & 0.417 & 0.728 & \textbf{0.385} & 0.769 &  0.648 \\
       ShaDDR & 0.217 & 0.636 & 0.220 & 0.522 & 0.673  \\
       Ours & \textbf{0.419} & \textbf{0.757} & 0.358 & \textbf{0.771} & \textbf{0.629}  \\
    \bottomrule
    \end{tabular}
    }
\end{minipage}
\end{table}

\section{More results on multi-category style mixing}

We show additional qualitative results on the chair and table style mixing in Figure \ref{fig:multi_category_ct_1} and ~\ref{fig:multi_category_ct_2}, and plant, building, cake and crystal in Figure \ref{fig:multi_category_pbcc_1} and \ref{fig:multi_category_pbcc_2}.

\section{More analysis of the generative capability}

\qm{
Our network design and losses (e.g., structure preservation) train the model to pick reasonable/plausible details to generate for each part regardless of its spatial location, as unreasonable/implausible details will lead to sub-optimal losses.
This contributes to the fact that most of our results exhibit transfers between structurally matched parts, which, under normal circumstances, would also represent parts that share similar spatial locations (e.g., chair back to back).

That said, our generative capability is certainly not constrained by the relative spatial positions of the parts.
Please see Fig.~\ref{fig:more_part_labels_ablation_results} for additional examples. 
Specifically, Figs.~\ref{fig:more_part_labels_ablation_results} (a-c) show that our model can generate armrests in various positions, even on just {\em one side\/} (b). Fig.~\ref{fig:more_part_labels_ablation_results} (d) shows a result of applying the style of tabletop to the chair back and (e) shows a chair back style detailized both in the middle of and in front of the seat, even between chair legs.

Even when part labels are removed from the input, our method is still able to generate reasonable geometry that respects both the style shape and the structure of the coarse voxel shape, as shown in Fig.~\ref{fig:more_part_labels_ablation_results} (f).

Note that changing style guidance from one to another \textit{within} the training style shapes does not require retraining. On the other hand, the model needs to be retrained to include new style shapes that are unseen during training.
}

\section{GUI demo}

\qm{After the user assigns styles to each region, our method takes less than a second ($\sim$0.3s) to generate style-mixing results}. We provide a video of our GUI demo powered by the network \textit{with} part labels version in the supplementary.

\input{figures_supp/more_part_labels_ablation_results}

\clearpage
\begin{figure*}
\begin{picture}(0, 500)
\centering
  \put(0, 0){\includegraphics[width=1.0\linewidth]{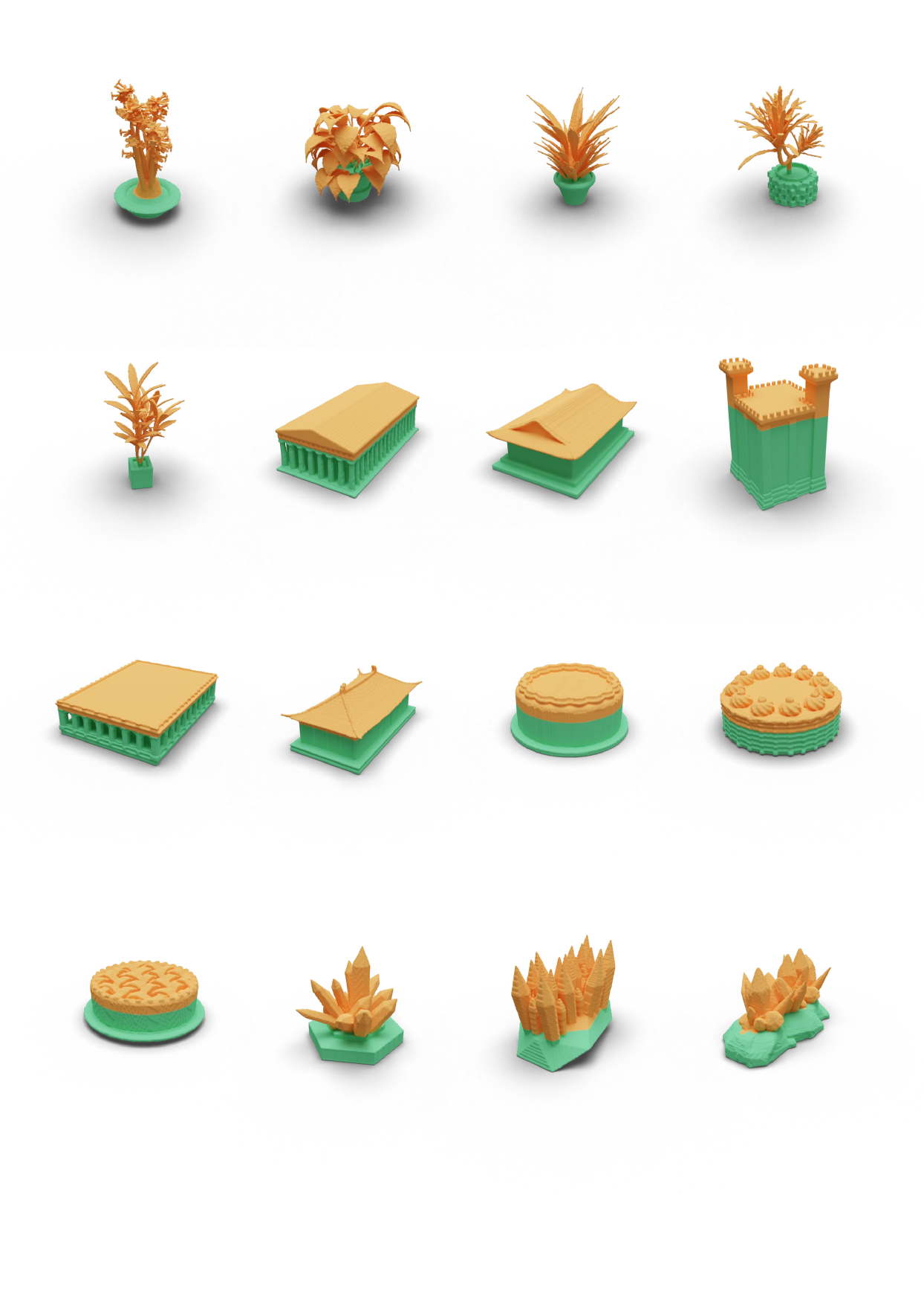}}
\end{picture}
  \caption{The segmented training style shapes of plant, building, cake and crystal categories. We segment the plant into two parts: pot and leaves, the building into two parts: main body and roof, the cake and crystal into two parts: bottom and top.}
  \label{fig:plant_building_cake_crystal_style_shapes}
\end{figure*}

\input{figures_supp/single_category_building_plant}

\input{figures_supp/multi_category_ct_1}
\input{figures_supp/multi_category_ct_2}

\input{figures_supp/multi_category_pbcc_1}
\input{figures_supp/multi_category_pbcc_2}

\end{document}